%% file: main.tex
\documentclass[10pt,twocolumn,letterpaper]{article}

\usepackage{cvpr}              % To produce the CAMERA-READY version
\input{preamble}
\definecolor{cvprblue}{rgb}{0.21,0.49,0.74}
\usepackage[pagebackref,breaklinks,colorlinks,allcolors=cvprblue]{hyperref}
\usepackage{hyperref}
\usepackage{url}
\setcitestyle{numbers}
\setcitestyle{square}
\usepackage{wrapfig}
\usepackage{xcolor}
\usepackage{color} 
\usepackage{listings}
\usepackage{amsmath,amsfonts,bm}
\usepackage[table,dvipsnames]{xcolor}
\usepackage{multirow}
\usepackage{collcell}
\usepackage{subcaption}
\usepackage{array}
\usepackage{arydshln}
\usepackage{algorithm}
\usepackage{algpseudocode}
\usepackage{algorithmicx}
\usepackage{listings}
\usepackage{enumitem}
\usepackage{etoolbox}
\usepackage[accsupp]{axessibility}
\usepackage{graphicx}
\usepackage{cancel}
\usepackage{threeparttable}
\usepackage{tikz} 
\usepackage{pifont}
\usepackage{bibunits}
\definecolor{codegreen}{RGB}{79,126,127}
\definecolor{codedefine}{RGB}{153,54,159}
\definecolor{codefunc}{RGB}{73,122,234}
\definecolor{codecall}{RGB}{73,122,234}
\definecolor{codepro}{RGB}{212,96,80}
\definecolor{codedim}{RGB}{89,152,195}
\definecolor{codeyellow}{RGB}{255, 215, 0}
\definecolor{mycodeblue}{RGB}{70, 130, 180}

\definecolor{best}{rgb}{1.0, 0.6, 0}
\definecolor{best2}{rgb}{1.0, 0.8, 0.6}
\newcommand{\irislogo}{%
    \raisebox{-0.2ex}{\includegraphics[height=1.2em]{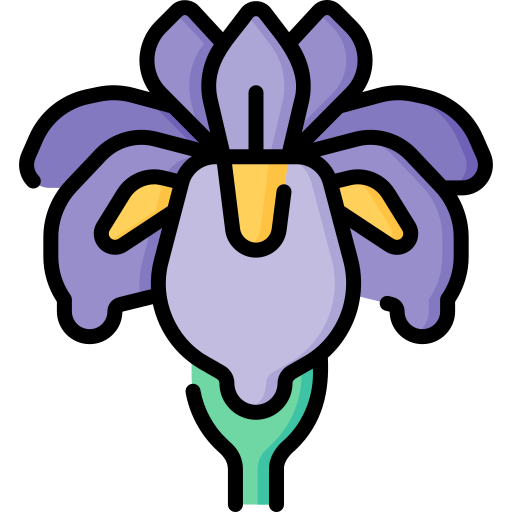}}%
}

\setlist[itemize]{leftmargin=16pt}

\definecolor{mygreen}{rgb}{0.20, 0.65, 0.27}
\definecolor{myred}{rgb}{0.65, 0.20, 0.27}

\makeatletter
\newcommand{\thickhline}{%
    \noalign {\ifnum 0=`}\fi \hrule height 1pt
    \futurelet \reserved@a \@xhline
}

\newcommand{\spthickhline}{%
    \noalign {\ifnum 0=`}\fi \hrule height 1.5pt
    \futurelet \reserved@a \@xhline
}
\newcommand{\sspthickhline}{%
    \noalign {\ifnum 0=`}\fi \hrule height 1.6pt
    \futurelet \reserved@a \@xhline
}
\definecolor{lightred}{RGB}{255,204,203}

\makeatletter
\def\@fnsymbol#1{%
  \ensuremath{%
    \ifcase#1\or
      \dagger\or
      \ddagger\or
      \mathsection\or
      \mathparagraph\or
      \|\or
      **\or
      \dagger\dagger\or
      \ddagger\ddagger
    \else
      \@ctrerr
    \fi}}
\makeatother

\def\paperID{XXXX} % *** Enter the Paper ID here
\def\confName{CVPR}
\def\confYear{2026}

\title{ \irislogo\ Iris: Bringing Real-World Priors into \\ Diffusion Model for Monocular Depth Estimation}

\author{
\begin{tabular}{@{}l@{}}
\bfseries
Xinhao Cai$^{1,2}$,~
Gensheng Pei$^{3}$,~
Zeren Sun$^{1,2}$,~
Yazhou Yao$^{1,2}$\thanks{Corresponding author.},~
Fumin Shen$^{4}$,~
Wenguan Wang$^{5}$\footnotemark[1]
\end{tabular}\\[4pt]
\small $^{1}$ Nanjing University of Science and Technology \quad \small $^{2}$ State Key Laboratory of Intelligent Manufacturing of Advanced Construction Machinery \\
\small $^{3}$ Department of Electrical and Computer Engineering, Sungkyunkwan University \\
\small $^{4}$ University of Electronic Science and Technology of China \quad \small $^{5}$ Zhejiang University  \\
\small \textcolor[HTML]{da3593}{https://github.com/NUST-Machine-Intelligence-Laboratory/Iris}
}

\begin{document}
\maketitle
\input{sec/0_abstract}    
\input{sec/1_intro}
\input{sec/2_related}

\input{sec/3_method}

\input{sec/4_exp}

\input{sec/5_conclusion}
% \input{sec/X_suppl.tex}
% \clearpage
% {
%     \small
%     \bibliographystyle{ieeenat_fullname}
%     \bibliography{main}
% }
{
    \small
    \bibliographystyle{ieeenat_fullname}
    \bibliography{main}
}
\clearpage
\input{sec/X_suppl.tex}


\end{document}

%% file: sec/0_abstract.tex
\begin{abstract}
In this paper, we propose \textbf{Iris}, a deterministic framework for Monocular Depth Estimation (MDE) that integrates real-world priors into the diffusion model.  Conventional feed-forward methods rely on massive training data, yet still miss details. Previous diffusion-based methods leverage rich generative priors yet struggle with synthetic-to-real domain transfer. Iris, in contrast, preserves fine details, generalizes strongly from synthetic to real scenes, and remains efficient with limited training data. To this end, we introduce a two-stage Priors-to-Geometry Deterministic (PGD) schedule: the prior stage uses Spectral-Gated Distillation (SGD) to transfer low-frequency real priors while leaving high-frequency details unconstrained, and the geometry stage applies Spectral-Gated Consistency (SGC) to enforce high-frequency fidelity while refining with synthetic ground truth. The two stages share weights and are executed with a high-to-low timestep schedule. Extensive experimental results confirm that Iris achieves significant improvements in MDE performance with strong in-the-wild generalization.
\end{abstract}

%% file: sec/1_intro.tex
\section{Introduction}
\label{sec:introduction}
As a fundamental task in computer vision, monocular depth estimation underlies a wide variety of emerging applications, such as 3D reconstruction~\citep{wang2023sparsenerf}, autonomous driving~\citep{hu2023planning}, and conditional image generation~\citep{zhang2023adding}. Accurate per-pixel depth estimation hinges on robust scene modeling, capturing both global layout and local geometry. While deep learning has driven substantial gains, depth estimation still struggles with accuracy, fine-detail fidelity, and generalization to diverse in-the-wild scenes. The dominant bottleneck is the \emph{training data}: \ding{172} real-world datasets provide imperfect supervision with inaccurate depth maps and poor preservation of fine details~\citep{yang2024depthv2}. \ding{173} synthetic datasets, despite offering perfect annotations, are modest in scale and suffer from a pronounced domain gap with respect to diverse real imagery.

Depth Anything V2 (DAv2)~\citep{yang2024depthv2} pushes the limits of conventional feed-forward depth estimators by scaling training data: a detail-preserving teacher model is trained on synthetic datasets, subsequently deployed to pseudo-label large unlabeled real-image corpora, and its supervision is then distilled into a student model. Despite strong cross-domain generalization, DAv2 has two key practical limitations: it relies on a prohibitive training scale that is hard to replicate, and it still underperforms on fine-grained detail and boundary precision, even with synthetic datasets. 

Beyond merely scaling training data, recent studies exploit diffusion priors for zero-shot monocular depth estimation. These studies demonstrate that text-to-image diffusion models such as Stable Diffusion~\citep{rombach2022high}, pretrained on billions of internet-scale image-text pairs~\citep{schuhmann2022laion}, provide powerful and comprehensive visual cues that can be repurposed to elevate per-pixel accuracy. When fine-tuned on limited synthetic data, diffusion-based models can reconstruct fine details and boundaries without large-scale real supervision. However, the performance of diffusion-based methods remains suboptimal relative to DAv2. Moreover, they often struggle with synthetic-to-real transfer, exhibiting limited generalization beyond the synthetic training domain.

These concerns raise a central question: \emph{with limited labeled data and compute, can we build a model that preserves fine-grained detail, generalizes strongly across domains, and achieves accuracy competitive with or surpassing models trained on enormous datasets?} To this end, we introduce \textbf{Iris}, a diffusion-based Priors-to-Geometry framework that integrates \emph{real-world priors} to jointly enhance cross-domain generalization and accuracy on real benchmarks while maintaining fidelity on fine details and boundaries, all within modest data and compute budgets. Iris converts the stochastic diffusion paradigm into a deterministic, feed-forward architecture tailored to dense prediction, eliminating iterative sampling and improving training and inference efficiency, as also observed in recent studies~\citep{garcia2025fine,xu2025matters}. 

\begin{figure}[t!]
    \centering
        \centerline{\includegraphics[width=0.98\linewidth]{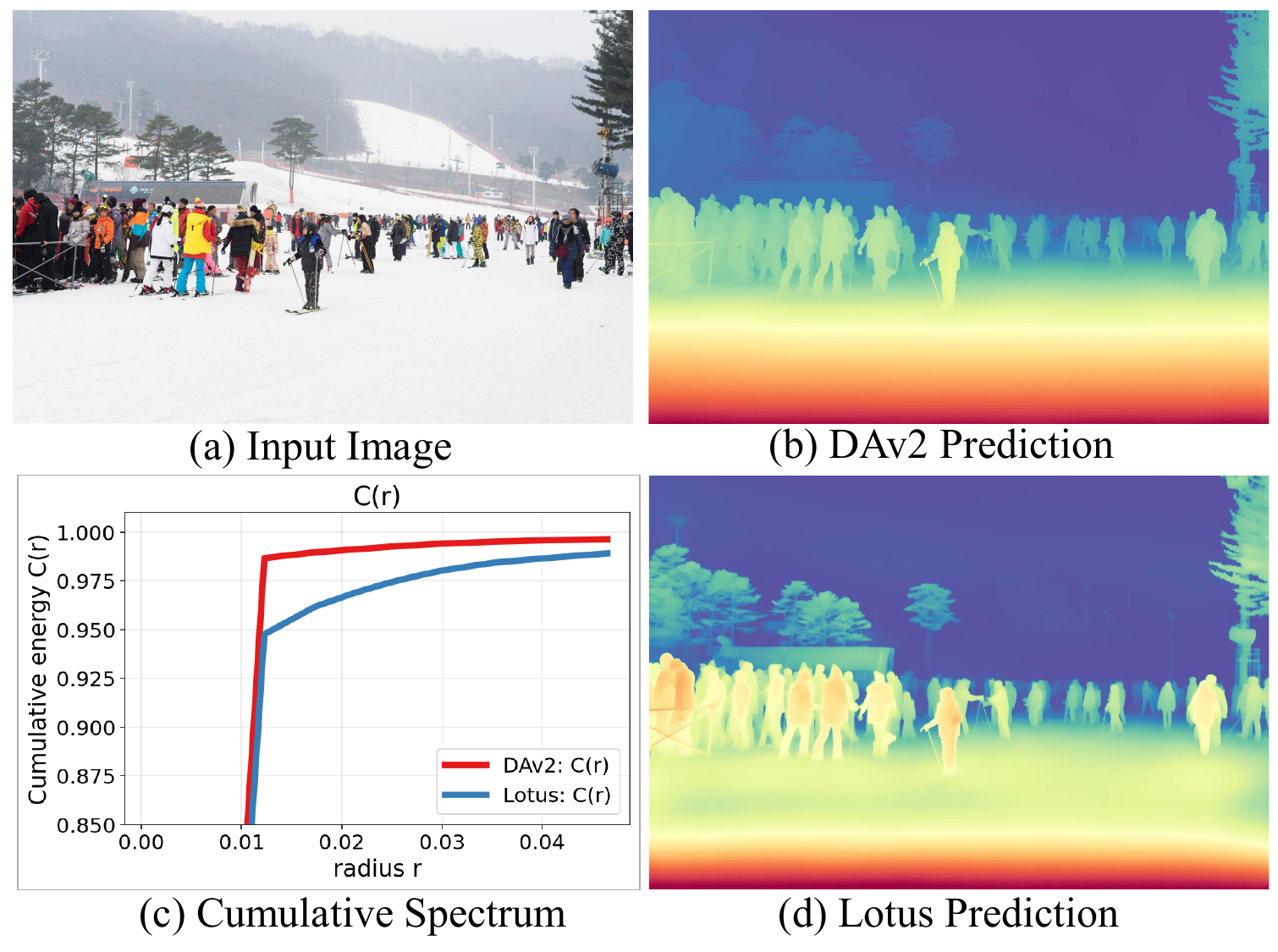}}
        \vspace{-10pt}
        \caption{\textbf{Comparison of DAv2 and diffusion-based method.} (a) Input. (b) DAv2~\citep{yang2024depthv2} yields accurate global layout and scale but smoother details. (d) The diffusion-based method (\textit{i.e.,} Lotus~\citep{he2025lotus}) preserves fine details and sharper boundaries. This complementarity motivates our Priors-to-Geometry Deterministic (\S\ref{sec: PGD}) framework; spectral disparity further motivates Spectral-Gated Distillation (\S\ref{sec: SGD}), which transfers reliable low-frequency real-image priors while deferring high-frequency details.}
        \label{fig:energy1}
\vspace{-10pt}
\end{figure}

To imbue Iris with real-world priors, we employ a teacher-student distillation framework where a depth estimator trained on real-world images supervises the diffusion model through prior distillation. In practice, this knowledge transfer process is non-trivial. We observe that directly using single-step deterministic perception~\citep{garcia2025fine,xu2025matters,he2025lotus} is ill-suited for teacher-student distillation. There is a \emph{frequency-reliability mismatch} (Fig.~\ref{fig:energy1}): teacher pseudo labels on real images are reliable for low-frequency structure (\textit{i.e.}, global layout) yet underspecify high-frequency content (\textit{i.e.}, fine details). Conversely, the supervision required to acquire high-frequency fidelity comes from synthetic datasets with precise ground truth. Training in a single pass forces the student to \emph{reconcile these opposing signals simultaneously}, causing gradient interference that degrades detail modeling and may imprint teacher-specific artifacts. 

\begin{figure}[t!]
    \centering
        \centerline{\includegraphics[width=0.98\linewidth]{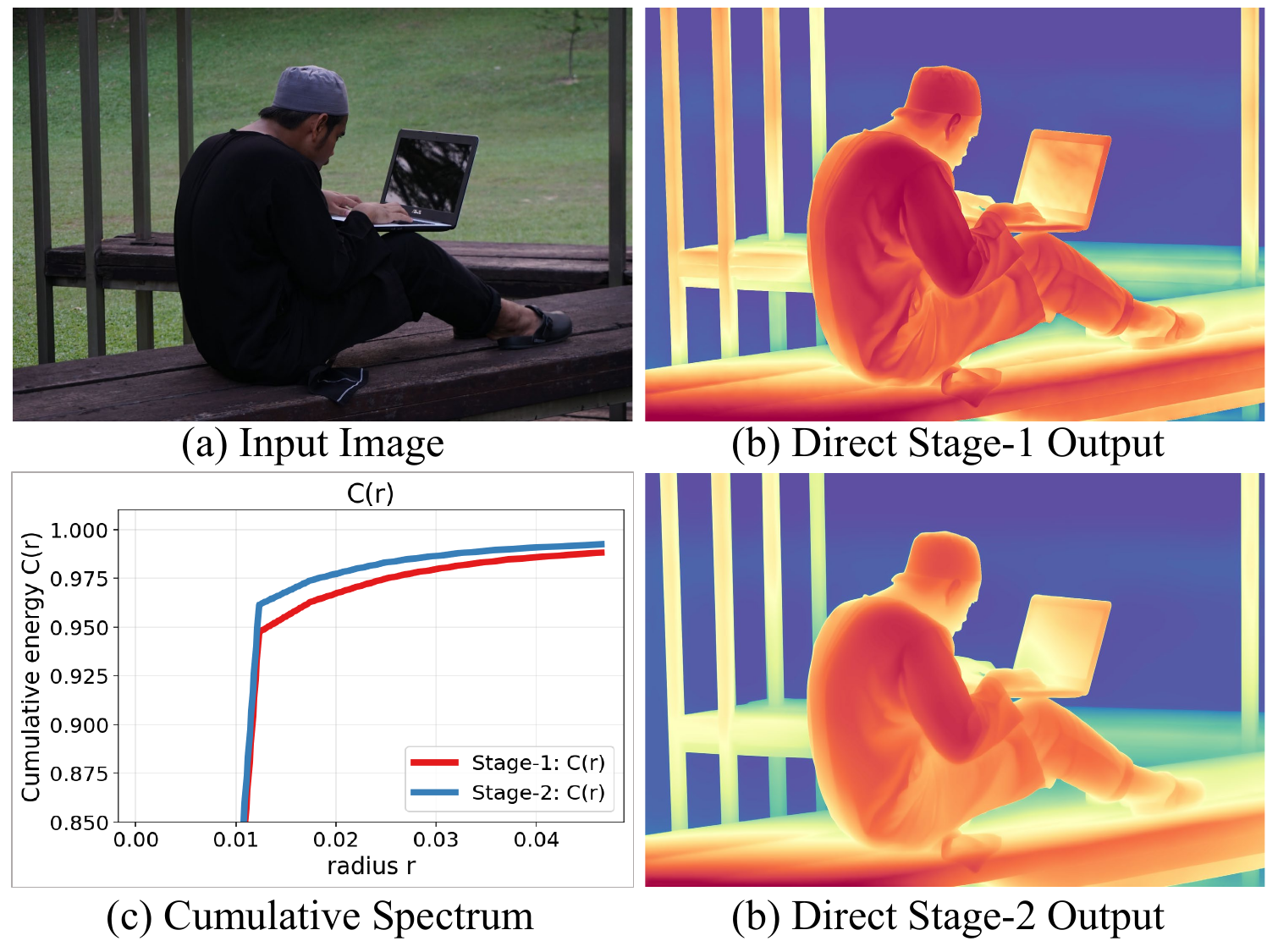}}
        \vspace{-10pt}
        \caption{\textbf{Comparison of direct stage-1 and stage-2 outputs.} (a) Input. (b) Unexpectedly, stage-1 operating at a high timestep with low-pass prior alignment produces crisp boundaries and richer textures. (d) The low-timestep stage-2 refined with synthetic ground truth yields smoother boundaries and more stable geometry. (c) Cumulative spectrum shows that stage-1 carries stronger high-frequency energy. These observations motivate using stage-1 as a high-frequency teacher via Spectral-Gated Consistency (\S\ref{sec: SGC}).}
        \label{fig:energy2}
\vspace{-8pt}
\end{figure}

In response, we propose the Priors-to-Geometry Deterministic (PGD) framework. In the first prior-alignment stage, the predictor operates at a high diffusion timestep corresponding to the low-SNR regime of the diffusion schedule, and a frozen real-image teacher supervises the predictor to calibrate global layout and metric scale while leaving high-frequency components largely unconstrained. In the subsequent geometry-refinement stage, the predictor switches to a low timestep in the high-SNR regime, and it is trained with synthetic ground truth to acquire high-frequency detail and precise geometry, including thin structures and boundaries.

Specifically, in the first prior-alignment stage, we introduce Spectral-Gated Distillation (SGD) to further address the \emph{frequency-reliability mismatch} problem. SGD learns a lightweight, differentiable low-pass gate in the Fourier domain that softly attenuates the teacher's high-frequency content while passing reliable low-band structure. The student is trained to match only the gated spectrum of the teacher, which transfers domain-robust cues for global layout and scale without imprinting teacher-specific high-frequency artifacts. High-frequency channels are intentionally left unconstrained in this stage and are learned in the subsequent refinement stage from synthetic ground truth. Surprisingly, in the first stage, we observe that the predictor often produces sharp boundaries and fine-scale textures (Fig.~\ref{fig:energy2}). This effect stems from low-pass alignment concentrating supervision on stable global structure, thereby enforcing steeper boundary transitions. Inspired by this phenomenon, we propose Spectral-Gated Consistency (SGC): a differentiable high-pass gate aligns stage-2 with stage-1 in the high-frequency band, while an auxiliary constraint suppresses over-activation in stage-1 to stabilize detail transfer.
The key contributions of this paper are as follows:
\begin{itemize}
    \item We present \textbf{Iris}, a deterministic diffusion-based framework that integrates real-world priors, delivering strong cross-domain generalization and real-world accuracy while preserving fine-detail and boundary fidelity, all within modest data and compute budgets.
    \item We introduce a Priors-to-Geometry Deterministic (PGD) framework, which consists of prior alignment at a high diffusion timestep followed by geometry refinement at a low timestep; this separation decouples prior transfer from reconstruction and mitigates gradient interference. The two stages share the same weights.
    \item We introduce Spectral-Gated Distillation (SGD) and Spectral-Gated Consistency (SGC). SGD distills low-frequency priors from a frozen real-image teacher via a lightweight low-pass gate; SGC aligns stage-2 to stage-1 in the high-frequency band using a differentiable high-pass gate with an over-activation constraint.
\end{itemize}

Extensive experimental results confirm that Iris establishes new benchmarks in zero-shot monocular depth estimation. Iris achieves the best overall performance among all methods. Compared with the previous diffusion-based SoTA methods, Iris exhibits strong cross-domain generalization and delivers consistent gains on all real-image benchmarks. Against approaches that rely on massive data such as DAv2, Iris attains leading accuracy on most datasets and excels in fine-detail and boundary fidelity, while maintaining a small training cost and ensuring reproducibility under resource constraints.

%% file: sec/2_related.tex
\section{Related Work}
\label{sec:related_work}

\noindent\textbf{Monocular Depth Estimation.}
Alongside broader advances in visual learning~\citep{deng2009imagenet, Yin2025, yin2025uncertainty, zhou2025unialign, CA2C, sheng2024foster}, monocular depth estimation has remained an important and long-standing topic in computer vision. Starting from CNN-based methods~\citep{eigen2014depth, fu2018deep, lee2019big, yuan2022neural}, early depth estimation methods focus on predicting relative depth on specific datasets. To build a depth estimator that generalizes to unseen data, subsequent efforts~\citep{li2018megadepth, yin2020diversedepth, ranftl2020towards, yin2023metric3d, hu2024metric3d} have expanded model capacity and scaled the size and diversity of the training data. Vision Transformer-based methods~\citep{eftekhar2021omnidata, ranftl2021vision} continue to advance performance. More recently, Depth Anything~\citep{yang2024depth, yang2024depthv2} series scales to millions of training images and demonstrates strong performance across diverse scenarios. Despite progress, conventional feed-forward monocular depth estimators remain constrained by noisy and imperfect real-image supervision, which hampers detail and boundary fidelity, and they typically require massive training scale to sustain performance.

\noindent\textbf{Text-to-Image Diffusion Models.} Diffusion probabilistic models~\citep{sohl2015deep} learn to reverse a forward Gaussian noising process and have progressed rapidly in both theory~\citep{dhariwal2021diffusion, ho2022classifier, kingma2021variational} and methodology~\citep{ho2020denoising, song2020denoising, song2020score}. In the realm of text-to-image (T2I) generation, methods~\citep{nichol2021glide, rombach2022high, cai2025cycle, cai2026unbiased} improve image quality and layout consistency. Further, Stable Diffusion (SD) model~\citep{rombach2022high}, trained on the large-scale image-text paired dataset LAION-5B~\citep{schuhmann2022laion}, performs diffusion in a VAE latent space, thereby compressing the generative process and substantially improving sampling efficiency and image quality. In this work, we retain the SD architecture and repurpose it for geometry prediction by explicitly leveraging its broad and encyclopedic visual priors.

\noindent\textbf{Diffusion-based Perception Models.} Beyond T2I generation, diffusion has rapidly emerged as a powerful backbone for dense predictive vision tasks, including optical flow estimation~\citep{saxena2023surprising, luo2024flowdiffuser}, open-vocabulary semantic segmentation~\citep{li2023open}, monocular depth estimation~\citep{ke2024repurposing, fu2024geowizard, zavadski2024primedepth, garcia2025fine, xu2025matters, he2025lotus, bai2025fiffdepth, xu2025pixel}, and surface-normal prediction~\citep{ye2024stablenormal, xu2025matters, he2025lotus}, with competitive accuracy and notable data-efficiency. As pioneers, Marigold~\citep{ke2024repurposing} and GeoWizard~\citep{fu2024geowizard} repurpose the standard diffusion formulation for dense prediction, demonstrating the promise of diffusion models in perception. Furthermore, a widely-noted observation in diffusion-based perception is that training solely on synthetic data can preserve accuracy and fine-grained details, owing to its inherently dense and complete annotations. Extending these efforts, GenPercept~\citep{xu2025matters} and StableNormal~\citep{ye2024stablenormal} investigate the feasibility of replacing multi-step diffusion with a single-step formulation. Subsequent works~\citep{garcia2025fine, xu2025matters, he2025lotus, bai2025fiffdepth} further develop the single-step diffusion pipeline, introducing optimizations tailored to predictive tasks. 

Unlike single-step formulations that collapse time conditioning and train only on synthetic data, we adopt a two-stage Priors-to-Geometry Deterministic (PGD) framework. Stage-1 operates at a high diffusion timestep (low SNR) and performs Spectral-Gated Distillation (SGD) on real images with pseudo labels, transferring real-world priors for global layout and scale while leaving high-frequency content unconstrained. Stage-2 switches to a low timestep (high SNR) and learns from synthetic ground truth for metric calibration and high-frequency geometry. This decoupling reduces gradient interference, improves robustness under synthetic-to-real domain shift, and preserves geometric fidelity. In addition, prior diffusion-based perception often trains a separate model per task; our formulation unifies depth and surface normals within a single deterministic diffusion model, sharing priors across tasks and improving scalability.

% \subsection{Monocular Depth Estimation}

%% file: sec/3_method.tex
\section{Methodology}
\label{sec:method}
In this section, we present Iris, a deterministic diffusion-based framework that integrates real-world priors into the diffusion model for MDE. We first introduce the overall two-stage Priors-to-Geometry Deterministic (PGD) framework in \S\ref{sec: PGD}. We then describe Spectral-Gated Distillation (SGD) and Spectral-Gated Consistency (SGC) in \S\ref{sec: SGD} and \S\ref{sec: SGC}. Finally, we formalize the training objective in \S\ref{sec: TO}.
\begin{figure*}[t!]
    \centering
        \centerline{\includegraphics[width=\textwidth]{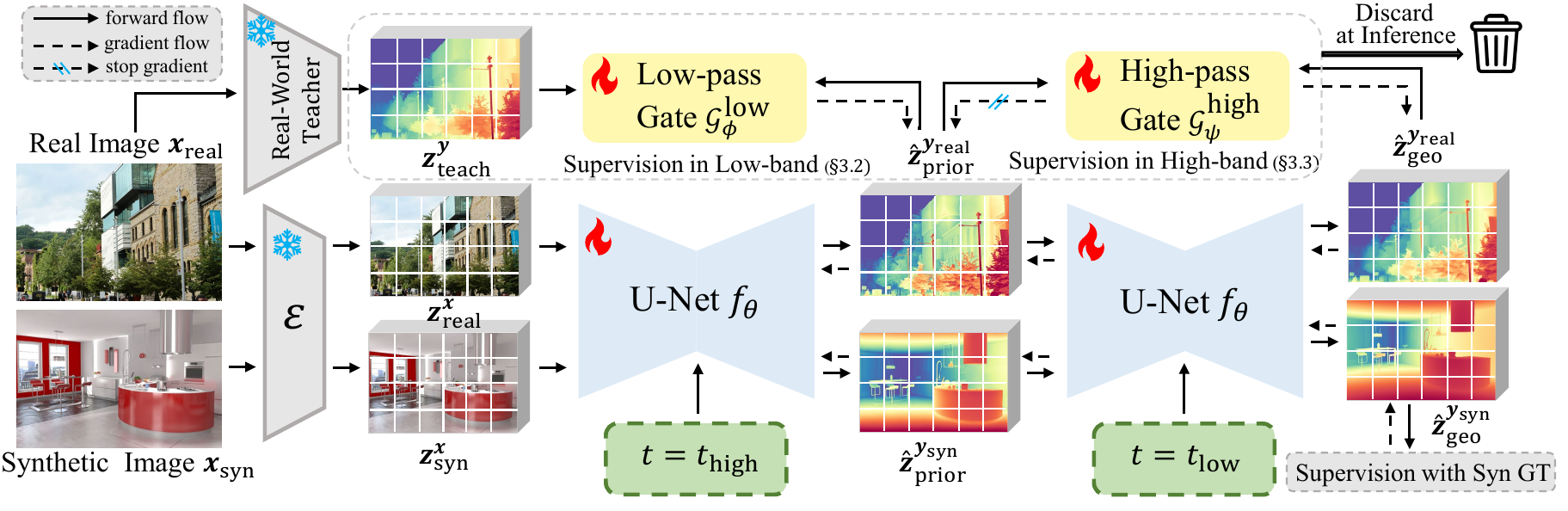}}
        \vspace{-10pt}
        \caption{\textbf{Iris overview}. Iris introduces a two-stage diffusion-based Priors-to-Geometry Deterministic framework that effectively injects real-world priors into the diffusion model. First prior stage injects real-world priors from a frozen teacher under a high-timestep state, while the second geometry stage refines metrically faithful predictions on synthetic supervision at a low-timestep state. In the prior stage, Spectral-Gated Distillation (\S\ref{sec: SGD}) uses a lightweight low-pass gate to filter noisy teacher predictions into stable low-frequency layout priors, whereas in the geometry stage, Spectral-Gated Consistency (\S\ref{sec: SGC}) applies a lightweight high-pass gate to transfer sharp boundaries and fine details from stage-1 to stage-2. \emph{The two U-Net blocks share weights.}
        Please refer to \S\ref{sec:method} for more details.}
        \label{fig:pipeline}
        \vspace{-10pt}
\end{figure*}

\subsection{Priors-to-Geometry Deterministic framework}
\label{sec: PGD}
Diffusion Models (DMs) map a source domain to a target domain.
Depth estimation, in turn, requires a mapping from images to structured labels, which aligns closely with this paradigm. However, the scarcity of expensive dense-annotated data often limits the precision of trained models. Recently, diffusion models pre-trained on internet-scale image corpora have shown great promise for knowledge transfer to dense prediction tasks~\citep{ke2024repurposing, yang2023diffusion, zhao2023unleashing}. Following this pipeline, we further build a diffusion-based framework.

We begin with the standard diffusion pipeline~\citep{ho2020denoising}, which progressively transforms Gaussian noise into a coherent image via iterative denoising. The forward process adds noise $\boldsymbol{\epsilon}\sim\mathcal{N}(0,I)$ to original image $\boldsymbol{x}_0$:
\begin{equation}
\label{equ:forward_process}
\boldsymbol{x}_t=\sqrt{\bar{\alpha}_t}\,\boldsymbol{x}_0+\sqrt{1-\bar{\alpha}_t}\,\boldsymbol{\epsilon},
\end{equation}
where $\bar{\alpha}_t=\prod_{s=1}^{t}\alpha_s$ and $\alpha_t=1-\beta_t$ follows a predefined variance schedule. At $t=T$, $\boldsymbol{x}_T$ approaches pure Gaussian noise. In the reverse process, a learnable denoiser $f_{\theta}$ (typically a U-Net~\citep{ronneberger2015u}) is trained to predict the added noise:
\begin{equation}
\label{equ:loss_dm}
\mathcal{L}_{\text{dm}}
=
\left\|
\boldsymbol{\epsilon}-f_\theta^{\boldsymbol{\epsilon}}\!\big(x_t,\,t\big)
\right\|_2^2 .
\end{equation}

In depth estimation, we adopt an image-conditioned diffusion formulation built on Stable Diffusion~\citep{rombach2022high}. An auto-encoder is used to map between RGB space and latent space. Given the image $\boldsymbol{x}$ and the dense annotation $\boldsymbol{y}$, the encoder maps them into the latent space: $\mathcal{E}(\boldsymbol{x})=\boldsymbol{z}^{\boldsymbol{x}}, \mathcal{E}(\boldsymbol{y})=\boldsymbol{z}^{\boldsymbol{y}}$. Additionally, inspired by previous works~\citep{xu2025matters,he2025lotus}, we directly adopt a deterministic pipeline and discard the multi-step diffusion mechanism. We employ the U-Net denoiser as the feed-forward network:
\begin{equation}
    \hat{\boldsymbol{z}}^{\boldsymbol{y}}=f_{\theta}\!\big(\boldsymbol{z}^{\boldsymbol{x}},\,t\big) .
\end{equation}
Since no stochastic noise is injected, this mapping is noise-independent and fully deterministic; the timestep $t$ serves only as a conditioning index of the diffusion state.

Previous diffusion-based perception models~\citep{ke2024repurposing,ye2024stablenormal,xu2025matters,he2025lotus,garcia2025fine} typically fine-tune on small-scale synthetic dense datasets with perfect annotations and achieve competitive performance with clear boundaries and details. However, this training regime often results in poor generalization to real images due to domain shift and rendering biases. 

To this end, we propose a Priors-to-Geometry Deterministic framework built on diffusion model that supervises a single shared-weight predictor under two diffusion states to bring real-image priors into metrically faithful geometry.

In the first stage, we operate the predictor at a high timestep $t_\text{high}$, corresponding to the low-SNR regime of the diffusion schedule. Although no noise is injected, conditioning on $t_\text{high}$ steers the predictor toward global layout and boundary structure while de-emphasizing fine textures. The first stage prior latent is given by:
\begin{equation}
\label{eq:stage1_mapping}
\hat{\boldsymbol{z}}^{\boldsymbol{y}}_{\text{prior}}
= f_{\theta}\!\big(\boldsymbol{z}^{\boldsymbol{x}},\, t_\text{high}\big).
\end{equation}
The prior-injection mechanism is detailed in \S\ref{sec: SGD}. While $\hat{\boldsymbol{z}}^{\boldsymbol{y}}_{\text{prior}}$ conveys global layout and sharp boundaries, it remains coarse and susceptible to pseudo label bias, motivating a second stage that refines geometry with synthetic supervision at a lower timestep $t_{\text{low}}$:
\begin{equation}
\label{eq:stage2_mapping}
\hat{\boldsymbol{z}}^{\boldsymbol{y}}_{\text{geo}}
= f_{\theta}\!\big(\hat{\boldsymbol{z}}^{\boldsymbol{y}}_{\text{prior}},\, t_{\text{low}}\big).
\end{equation}
% \begin{equation}
% \boldsymbol{z}_t^{\boldsymbol{y}}=\sqrt{\bar{\alpha}_t}\,\boldsymbol{z}^{\boldsymbol{y}}+\sqrt{1-\bar{\alpha}_t}\,\boldsymbol{\epsilon}
% \end{equation}

\subsection{Spectral-Gated Distillation}
\label{sec: SGD}
\begin{figure}[t!]
    \centering
        \centerline{\includegraphics[width=0.98\linewidth]{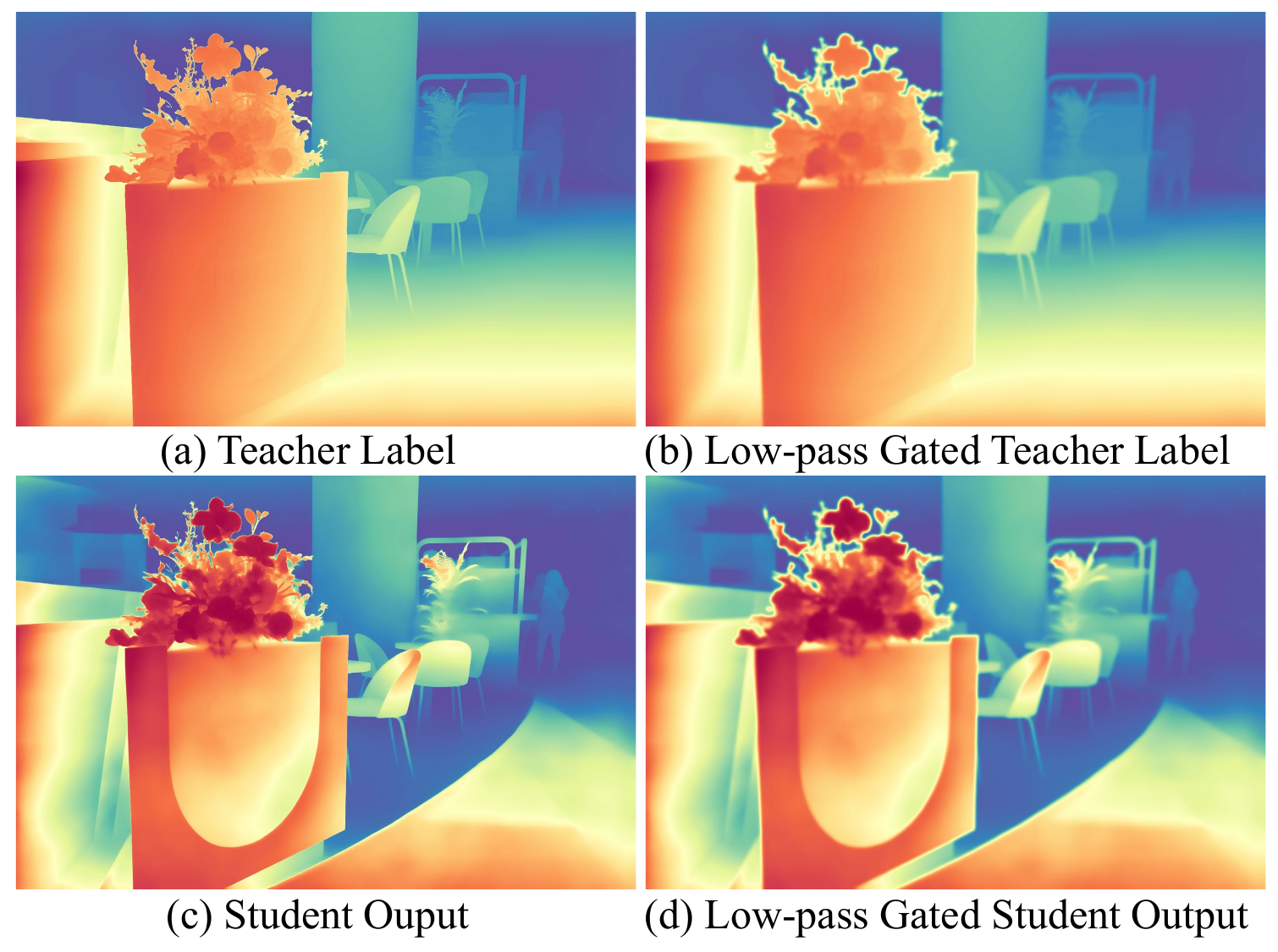}}
        \vspace{-8pt}
        \caption{\textbf{Visualization of Spectral-Gated Distillation.} SGD aligns teacher and student in the low-frequency band, injecting real-world priors for layout and scale, suppressing high-frequency artifacts, and leaving high-frequency components unconstrained for next-stage refinement. See \S\ref{sec: SGD} for more details.}
        \label{fig:distill1}
\vspace{-10pt}
\end{figure}

To transfer real-world priors in Eq.\!~(\ref{eq:stage1_mapping}), we first obtain pseudo labels on real images from an off-the-shelf teacher model pretrained on large-scale data, which offers broad coverage of scenes from the real world. However, pseudo labels from the teacher are reliable mainly in low-frequency bands (\textit{i.e.,} layout and object extents), whereas diffusion-based perception outputs carry pronounced high-frequency information (\textit{i.e.,} fine textures and sharper boundaries), as shown in Fig.~\ref{fig:distill1}. Therefore, naively regressing to pseudo labels can suppress high-frequency structures. We therefore introduce Spectral-Gated Distillation, which learns a lightweight Fourier low-pass gate and distills only low-band components, leaving high-frequency details intact.

Concretely, we define a lightweight learnable low-pass gate $\mathcal{G}_{\phi}$ in latent space with only three parameters:
\begin{equation}
\label{eq:sg_lp}
\begin{aligned}
\mathcal{G}^{\text{low}}_{\phi}(\boldsymbol{z})
&= \boldsymbol{z}
+ s\!\left(\mathcal{F}^{-1}\!\big(M_{\phi}\odot\mathcal{F}(\boldsymbol{z})\big)-\boldsymbol{z}\right) \\
M_{\phi}(\omega)
&= \mathrm{Sigmoid}\!\big(\beta\,(\kappa-\|\omega\|_2)\big),
\end{aligned}
\end{equation}
where $\phi=\{\kappa,\beta,s\}$ are the learnable cutoff, slope, and residual strength, $\mathcal{F}$/$\mathcal{F}^{-1}$ denote FFT/iFFT, and $\odot$ is elementwise product. Let $P$ be the teacher model; for a real image $\boldsymbol{x}\sim\mathcal{D}_{\text{real}}$, define its pseudo label $\boldsymbol{y}_{\text{teach}}(\boldsymbol{x})=P(\boldsymbol{x})$ and the corresponding latent $\boldsymbol{z}^{\boldsymbol{y}}_{\text{teach}}(\boldsymbol{x})=\mathcal{E}\big(\boldsymbol{y}_{\text{teach}}(\boldsymbol{x})\big)$. The spectral-gated distillation loss is:
\begin{equation}
\label{eq:sg_loss}
\mathcal{L}_{\text{sgd}}
=\mathbb{E}_{\boldsymbol{x}\sim\mathcal{D}_{\text{real}}}
\left\|
\mathcal{G}^{\text{low}}_{\phi}\!(\hat{\boldsymbol{z}}^{\boldsymbol{y}}_{\text{prior}}(\boldsymbol{x}))
-\mathcal{G}^{\text{low}}_{\phi}\!(\boldsymbol{z}^{\boldsymbol{y}}_{\text{teach}}(\boldsymbol{x}))
\right\|_2^2 .
\end{equation}
SGD acts as a parameter-light, data-adaptive filter that converts noisy teacher supervision into a stable low-band signal, enabling efficient transfer of real-world priors without erasing boundary sharpness of the diffusion model output.

\subsection{Spectral-Gated Consistency}
\label{sec: SGC}
\begin{figure}[t!]
    \centering
        \centerline{\includegraphics[width=0.98\linewidth]{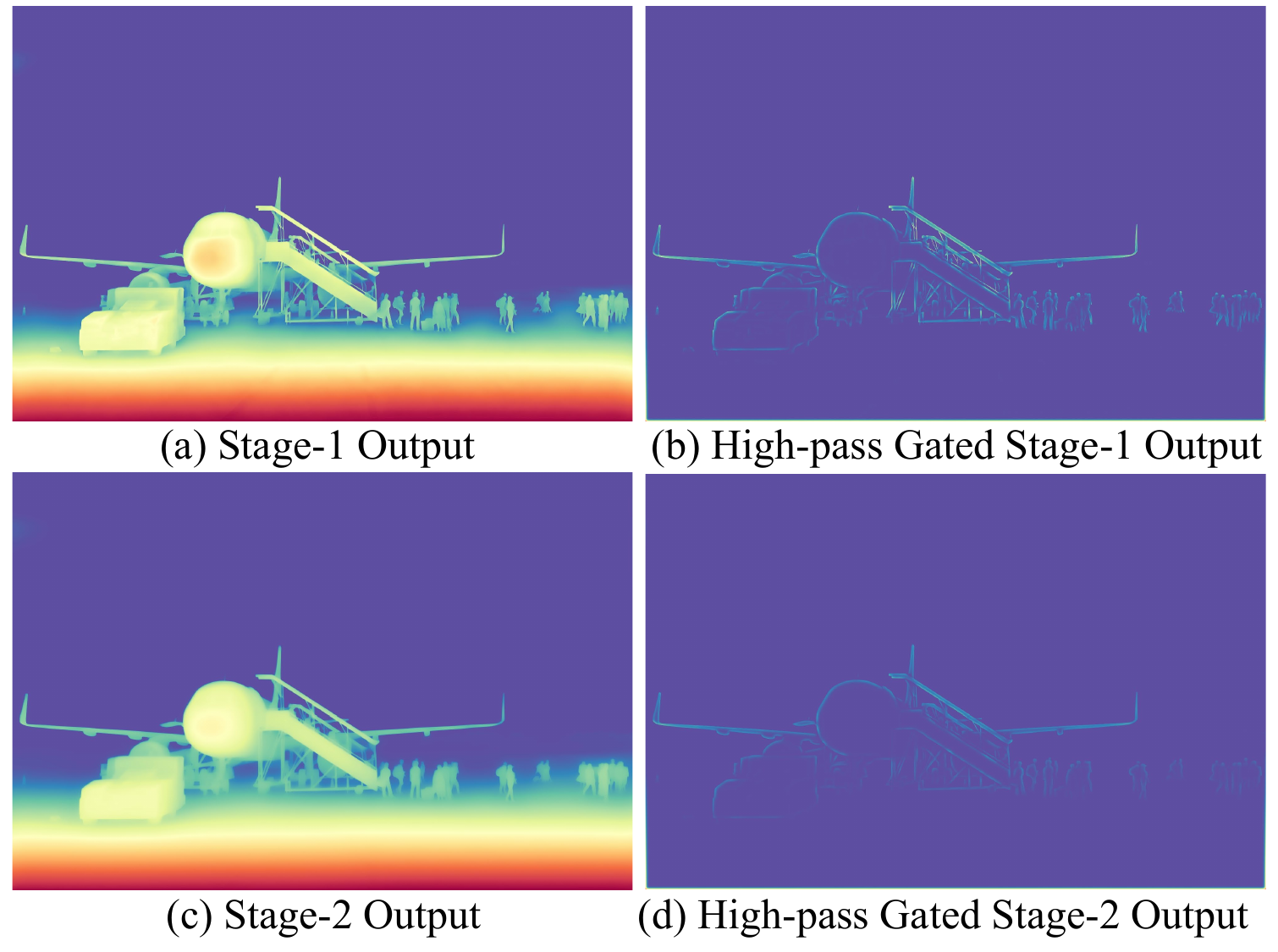}}
        \vspace{-8pt}
        \caption{\textbf{Visualization of Spectral-Gated Consistency.} Stage-1 naturally yields crisp detail and boundary cues. To leverage these internal cues, SGC encourages agreement between stages in the high-frequency band. See \S\ref{sec: SGC} for more details.}
        \label{fig:distill2}
\vspace{-10pt}
\end{figure}

We observe that the stage-1 predictor, although trained with low-pass alignment to pseudo labels, often produces sharper boundaries and fine-scale structures, as shown in Fig.~\ref{fig:distill2}. Concentrating supervision on low-frequency layout reduces conflicting high-band signals from the teacher and implicitly favors steeper transitions at semantic edges. This makes stage-1 a useful source of high-frequency guidance. Inspired by this, we encourage stage-2 to inherit high-frequency cues from stage-1 while keeping stage-1 stable.

Reusing the mask in Eq.~(\ref{eq:sg_lp}), define the complementary high-pass mask
$\overline{M}_{\phi}(\omega)=1-M_{\phi}(\omega)$ and a lightweight high-pass gate:
\begin{equation}
\label{eq:sgc_hp}
\mathcal{G}^{\text{high}}_{\psi}(\boldsymbol{z})
= \boldsymbol{z}
+ s_h\!\left(\mathcal{F}^{-1}\!\big(\overline{M}_{\phi}\odot\mathcal{F}(\boldsymbol{z})\big)-\boldsymbol{z}\right),
\end{equation}
where $\psi=\{\kappa_h,\beta_h,s_h\}$ are independent parameters different from $\phi$ in Eq. (\ref{eq:sg_loss}). The Spectral-Gated Consistency loss aligns stage-2 to stage-1 only in the high-frequency band, with an explicit stop-gradient on the teacher:
\begin{equation}
\label{eq:sgc_loss_match}
\mathcal{L}_{\text{sgc}}
=\mathbb{E}_{\boldsymbol{x}\sim\mathcal{D}_{\text{real}}}
\left\|
\mathcal{G}^{\text{high}}_{\psi}\!(\hat{\boldsymbol{z}}^{\boldsymbol{y}}_{\text{geo}}(\boldsymbol{x}))
-\mathrm{sg}\!\left[\mathcal{G}^{\text{high}}_{\psi}\!(\hat{\boldsymbol{z}}^{\boldsymbol{y}}_{\text{prior}}(\boldsymbol{x}))\right]
\right\|_2^2 ,
\end{equation}
where $\mathrm{sg}[\cdot]$ denotes stop-gradient.

\subsection{Training Objective}
\label{sec: TO}
For depth estimation, the final loss is:
\begin{equation}
\label{eq:dense}
\mathcal{L}_{\text{depth}}
=\mathbb{E}_{\boldsymbol{x}\sim\mathcal{D}_{\text{syn}}}
\left\|
\hat{\boldsymbol{z}}^{\boldsymbol{y}}_{\text{geo}}
-\boldsymbol{z}^{\boldsymbol{y}}
\right\|_2^2 + \alpha \mathcal{L}_{\text{sgd}} + \beta \mathcal{L}_{\text{sgc}},
\end{equation}
where $\hat{\boldsymbol{z}}^{\boldsymbol{y}}_{\text{geo}}$ is obtained via Eqs. (\ref{eq:stage1_mapping}-\ref{eq:stage2_mapping}) and therefore depends on $\boldsymbol{x}$, and $\beta$ controls the stage-1 over-activation constraint.
However, when fine-tuning SD, the catastrophic forgetting problem~\citep{zhai2023investigating} arises: optimizing only this dense-prediction loss tends to wash out high-frequency details and yields over-smoothed outputs, eroding the fine-grained modeling capacity inherited from text-to-image generation. 

\setlength{\tabcolsep}{2pt}
\renewcommand{\arraystretch}{1.05}

\begin{table*}[t]
\vspace{-3mm}
\caption{\textbf{Quantitative comparison on zero-shot affine-invariant depth estimation}. The upper section lists conventional discriminative methods, and the lower section lists diffusion-based methods. The best performances are \textbf{bolded}. \emph{All Avg Ranking} averages the per-metric ranks over all methods, while \emph{Group Avg Ranking} averages ranks computed within each section. $^\star$ denotes methods relying on pre-trained Stable Diffusion~\citep{rombach2022high}. \emph{Note that Iris is trained on 59K synthetic images and 100K real images with pseudo labels generated by DAv2}~\citep{yang2024depthv2}.Compared with deterministic feed-forward models trained on massive real-image corpora (e.g., the DA~\citep{yang2024depth,yang2024depthv2} family), Iris retains a clear advantage in training data efficiency and achieves competitive average performance. Relative to prior diffusion-based methods, Iris further delivers considerable performance gains. Please refer to \S\ref{sec:quan_result} for more details.}
\vspace{-6pt}

\resizebox{\textwidth}{!}{%
\begin{tabular}{l|c|cc|cc|cc|cc|cc|c|c}
\hline
\rowcolor[rgb]{0.92,0.92,0.92}
\rule{0pt}{2.6ex}\raisebox{-1.2ex}[2.6ex][1.3ex]{Method} & Training
& \multicolumn{2}{c|}{KITTI (Outdoor)}
& \multicolumn{2}{c|}{NYUv2 (Indoor)}
& \multicolumn{2}{c|}{ETH3D (Various)}
& \multicolumn{2}{c|}{ScanNet (Indoor)}
& \multicolumn{2}{c|}{DIODE (Various)}
& All Avg & Group Avg \\
\rowcolor[rgb]{0.92,0.92,0.92}
\rule{0pt}{2.4ex}
& Data$\downarrow$
& AbsRel$\downarrow$ & $\delta_1\uparrow$
& AbsRel$\downarrow$ & $\delta_1\uparrow$
& AbsRel$\downarrow$ & $\delta_1\uparrow$
& AbsRel$\downarrow$ & $\delta_1\uparrow$
& AbsRel$\downarrow$ & $\delta_1\uparrow$
& Ranking & Ranking \\
\hline

DiverseDepth~\citep{yin2020diversedepth}
& 320K
& 19.0 & 70.4
& 11.7 & 87.5
& 22.8 & 69.4
& 10.9 & 88.2
& 37.6 & 63.1
& 15.6 & 7.6 \\

MiDaS~\citep{ranftl2020towards}
& 2M
& 23.6 & 63.0
& 11.1 & 88.5
& 18.4 & 75.2
& 12.1 & 84.6
& 33.2 & 71.5
& 15.2 & 7.3 \\

LeRes~\citep{yin2021learning}
& 354K
& 14.9 & 78.4
& 9.0 & 91.6
& 17.1 & 77.7
& 9.1 & 91.7
& 27.1 & 76.6
& 12.3 & 5.3 \\

Omnidata~\citep{eftekhar2021omnidata}
& 12.2M
& 14.9 & 83.5
& 7.4 & 94.5
& 16.6 & 77.8
& 7.5 & 93.6
& 33.9 & 74.2
& 12.2 & 4.8 \\

DPT~\citep{ranftl2021vision}
& 1.4M
& 10.0 & 90.1
& 9.8 & 90.3
& \textbf{7.8} & \textbf{94.6}
& 8.2 & 93.4
& \textbf{18.2} & 75.8
& 9.9 & 3.5 \\

HDN~\citep{zhang2022hierarchical}
& 300K
& 11.5 & 86.7
& 6.9 & 94.8
& 12.1 & 83.3
& 8.0 & 93.9
& 24.6 & \textbf{78.0}
& 9.6 & 3.0 \\

DepthAnything~\citep{yang2024depth}
& 62.6M
& 7.6 & \textbf{94.7}
& \textbf{4.3} & \textbf{98.1}
& 12.7 & 88.2
& 4.3 & \textbf{98.1}
& 26.0 & 75.9
& 4.3 & \textbf{1.9} \\

DepthAnything V2~\citep{yang2024depthv2}
& 62.6M
& \textbf{7.4} & 94.6
& 4.5 & 97.9
& 13.1 & 86.5
& \textbf{4.2} & 97.8
& 26.5 & 73.4
& 5.4 & 2.7 \\

\hline

Diffusion-E2E-FT$^{\star}$~\citep{garcia2025fine}
& 74K
& 9.6 & 92.1
& 5.4 & 96.5
& 6.4 & 95.9
& 5.8 & 96.5
& 30.3 & \textbf{77.6}
& 5.6 & 4.0 \\

GeoWizard$^{\star}$~\citep{fu2024geowizard}
& 280K
& 14.4 & 82.0
& 5.6 & 96.3
& 6.6 & 95.8
& 6.4 & 95.0
& 33.5 & 72.3
& 9.9 & 7.1 \\

Marigold$_{\text{(LCM)}}$$^{\star}$~\citep{ke2024repurposing}
& 74K
& 9.8 & 91.8
& 6.1 & 95.8
& 6.8 & 95.6
& 6.9 & 94.6
& 30.7 & 77.5
& 8.5 & 6.7 \\

Marigold$^{\star}$~\citep{ke2024repurposing}
& 74K
& 9.9 & 91.6
& 5.5 & 96.4
& 6.5 & 95.9
& 6.4 & 95.2
& 30.8 & 77.3
& 7.4 & 5.8 \\

Lotus-D$^{\star}$~\citep{he2025lotus}
& 59K
& 8.1 & 93.1
& 5.1 & 97.2
& 6.1 & 97.0
& 5.5 & 96.5
& 22.8 & 73.8
& 4.4 & 2.8 \\

Lotus-G$^{\star}$~\citep{he2025lotus}
& 59K
& 8.5 & 92.2
& 5.4 & 96.8
& 5.9 & 97.0
& 5.9 & 95.7
& 22.9 & 72.9
& 5.6 & 3.8 \\

GenPercept$^{\star}$~\citep{xu2025matters}
& 90K
& 7.8 & 93.5
& 5.9 & 96.7
& 9.4 & 96.1
& 6.4 & 96.1
& \textbf{22.8} & 74.0
& 6.1 & 4.4 \\

\rowcolor[rgb]{0.92,0.92,0.92}
\rule{0pt}{2.4ex} Iris (\textbf{ours})$^{\star}$
& 59K + 100K
& \textbf{7.2} & \textbf{94.5}
& \textbf{4.9} & \textbf{97.4}
& \textbf{5.5} & \textbf{97.6}
& \textbf{5.0} & \textbf{97.1}
& 24.3 & 74.3
& \textbf{3.1} & \textbf{1.6} \\
\hline
\end{tabular}%
}
\label{tab:depth}
\vspace{-10pt}
\end{table*}

Inspired by~\citet{he2025lotus}, we introduce an auxiliary image reconstruction constraint to retain the fine-detail modeling capacity of the text-to-image backbone. Concretely, following the two-stage Priors-to-Geometry procedure, we activate the task switcher with \(s_x\) and reconstruct the input image for both real and synthetic samples via \(\hat{\boldsymbol{z}}^{\boldsymbol{x}}_{\text{prior}}=f_{\theta}\!\big(\boldsymbol{z}^{\boldsymbol{x}},\,t_{\text{high}},\,s_x\big)\) and \(\hat{\boldsymbol{z}}^{\boldsymbol{x}}_{\text{geo}}=f_{\theta}\!\big(\hat{\boldsymbol{z}}^{\boldsymbol{x}}_{\text{prior}},\,t_{\text{low}},\,s_x\big)\), followed by the reconstruction loss:
\begin{equation}
\label{eq:reconstruct}
\mathcal{L}_{\text{recon}}
=\mathbb{E}_{\boldsymbol{x}\sim\mathcal{D}_{\text{syn}}, \mathcal{D}_{\text{real}}}
\left\|
\hat{\boldsymbol{z}}^{\boldsymbol{x}}_{\text{geo}}
-\boldsymbol{z}^{\boldsymbol{x}}
\right\|_2^2 .
\end{equation}
Note that real-image supervision occurs at stage~1 (\S\ref{sec: SGD}), whereas reconstruction is performed only at stage~2 to leverage the low-timestep regime, and maintain consistency between synthetic and real images. We observe that adding a reconstruction term on real images at stage-2 improves detail retention on real-world scenes. The final loss is:
\begin{equation}
\label{eq:final_loss}
\mathcal{L}=\mathcal{L}_{\text{depth}}+\gamma\,\mathcal{L}_{\text{recon}}.
\end{equation}

%% file: sec/4_exp.tex
\section{Experiments}

\subsection{Experimental Setup}
\noindent\textbf{Training Datasets.} 
Our model is trained on two synthetic and one real-world dataset covering indoor and outdoor scenes:
\begin{itemize}
    \item \textbf{Hypersim}~\citep{roberts2021hypersim} is a photorealistic indoor dataset with 461 scenes. We use the official training split. Following Lotus~\citep{he2025lotus}, we remove incomplete entries and retain about 39K samples for training. All images are resized to 576 $\times$ 768 before training.
    \item \textbf{Virtual KITTI}~\citep{cabon2020virtual} is a synthetic urban driving dataset with five scenes under diverse imaging and weather conditions. The training set comprises four scenes and approximately 20K samples. All images are cropped to 352 $\times$ 1216, with the far plane set to 80$m$. 
    \item \textbf{SA-1B}~\citep{kirillov2023segment} is a large-scale real-world dataset introduced in Segment Anything, comprising 11M real images and 1.1B masks across diverse scenarios. We leverage only \textbf{100K} real images from SA-1B and generate pseudo labels using Depth Anything V2~\citep{yang2024depthv2}. All images are resized to 576 $\times$ 768 before training.
\end{itemize}
For each batch, we select one of the three datasets with fixed probabilities—Hypersim 60\%, Virtual KITTI 10\%, and SA-1B 30\%, and then draw all samples from the selected dataset.

\noindent\textbf{Evaluation Datasets.} 
\ding{172} For zero-shot affine-invariant depth estimation, we evaluate our model on five real datasets, including NYUv2~\citep{silberman2012indoor}, ScanNet~\citep{dai2017scannet}, KITTI~\citep{geiger2013vision}, ETH3D~\citep{schops2017multi}, and DIODE~\citep{vasiljevic2019diode}.

\noindent\textbf{Evaluation Metrics.} \ding{172} For zero-shot affine-invariant depth estimation, the accuracy of the aligned predictions is assessed by the \emph{absolute mean relative error} (AbsRel), \emph{i.e.}, $\frac{1}{M}\sum_{i=1}^M |a_i-d_i|/d_i$, where $M$ is the total number of pixels, $a_i$ is the predicted depth map and $d_i$ represents the ground truth. We also measure $\delta_1$, defined as the proportion of pixels satisfying $\text{max}(a_i/d_i, d_i/a_i)<1.25$.

\begin{figure*}[t]
    \centering
    \includegraphics[width=\textwidth]{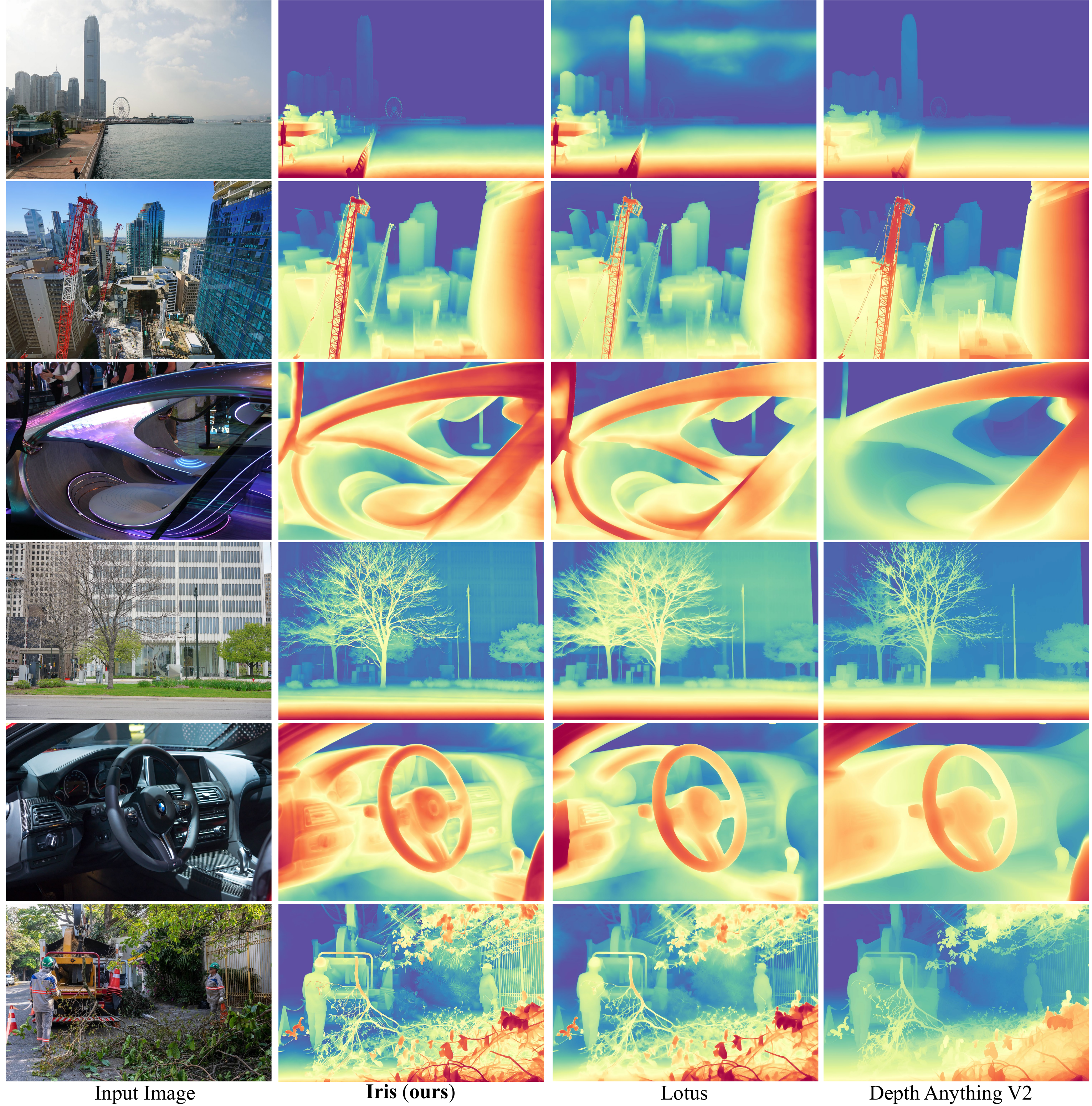}
    \vspace{-22pt}
    \caption{\textbf{Qualitative comparison on diverse scenes}. Iris demonstrates consistent cross-scene generalization and accurate fine-detail modeling. See \S\ref{sec:qual_result} for details. For more qualitative results, please refer to the supplementary material.}
    \label{fig:vis1}
    \vspace{-10pt}
\end{figure*}

\noindent\textbf{Implementation Details.} Iris is built on Stable Diffusion V2~\citep{rombach2022high} and omits text conditioning. For depth estimation, we predict in disparity space, \textit{i.e.}, $d = 1/d'$, where $d$ represents the values in disparity space and $d'$ denotes the true depth. All depth ground-truth maps are normalized to the $[-1,1]$ to align with the VAE's original input range. We employ Depth Anything V2-Large~\citep{yang2024depthv2} as the teacher model for distilling real-world priors. We first sample 100K real images from SA-1B~\citep{kirillov2023segment} and resize them to $1204^{2}$ resolution. The images are then fed to the teacher model to obtain zero-shot and affine-invariant disparity maps.
During training, the first timestep is fixed at $t=1000$ for real-world priors distillation and the second is fixed at $t=500$ for precise synthetic data supervision. We utilize the standard Adam optimizer with a learning rate of $7.5\times 10^{-6}$. $\alpha$ and $\gamma$ are both set to 1, and $\beta$ is set to 0.1 to add an over-activation constraint. Experiments are conducted on 4 NVIDIA A100 40GB GPUs, using a total batch size of 32.

\subsection{Quantitative Results}
\label{sec:quan_result}
As shown in Table~\ref{tab:depth}, Iris delivers consistently strong zero-shot affine-invariant depth estimation across all benchmarks and \emph{ranks first} in both \emph{All Avg Ranking} and \emph{Group Avg Ranking}. Within diffusion-based methods, Iris achieves the best accuracy on most datasets, clearly improving over previous Marigold~\citep{ke2024repurposing}, and the Lotus~\citep{he2025lotus} variants. Compared with deterministic feed-forward models trained on massive real-image corpora (\textit{e.g.,} the Depth Anything~\citep{yang2024depth,yang2024depthv2} family with 62.6M images), Iris is trained on only 59K synthetic images plus 100K real images with pseudo labels, yet it remains highly competitive and exhibits the strongest overall performance across all 16 methods. These results indicate that Iris effectively distills real-image priors while preserving strong geometric fidelity, leading to data-efficient and robust cross-dataset generalization.

\begin{figure}[t]
    \centering
        \centerline{\includegraphics[width=0.98\linewidth]{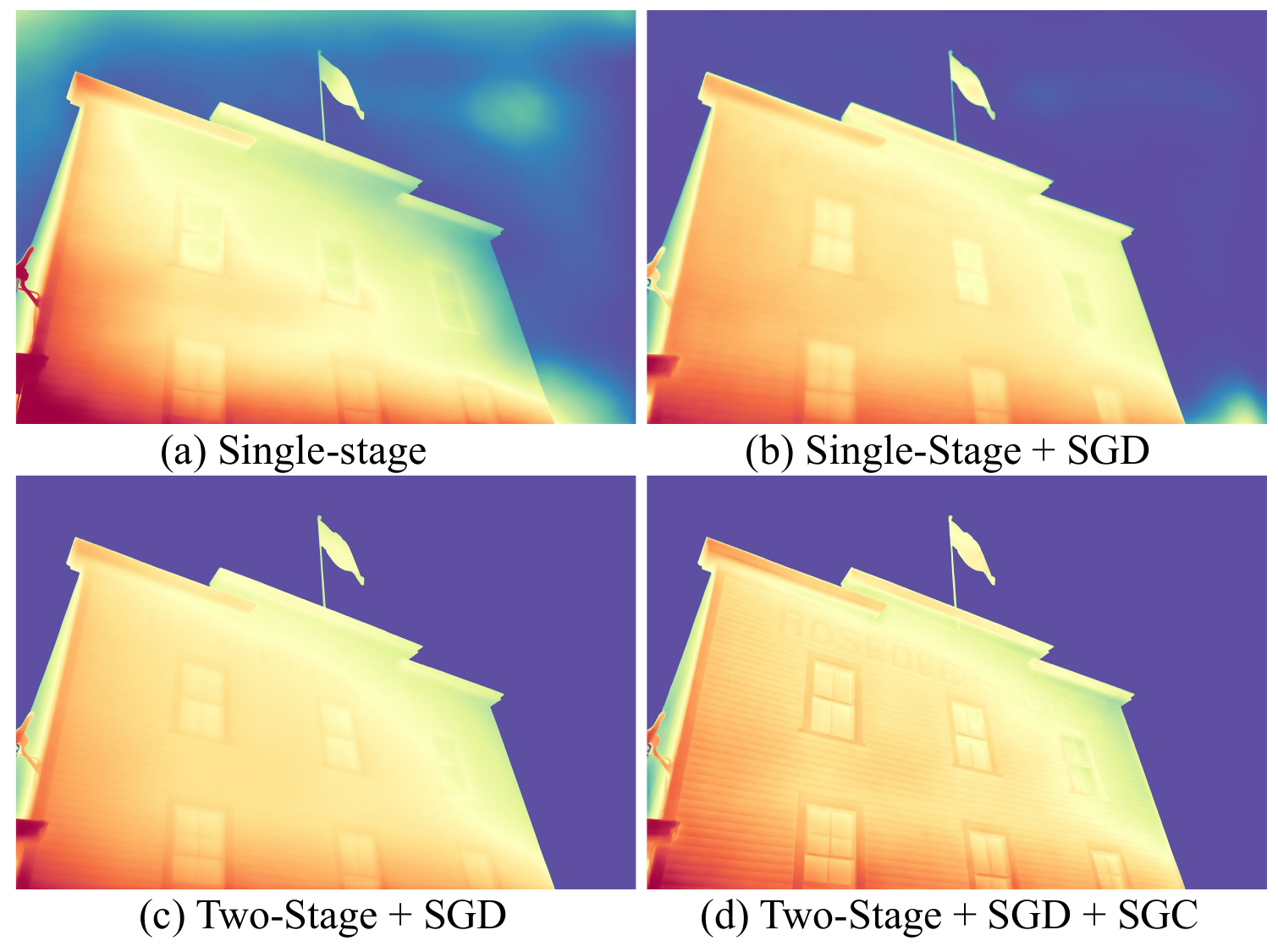}}
        \vspace{-12pt}
        \caption{\textbf{Visualization of ablation studies.} Note that two-stage denotes the Prior-to-Geometry pipeline. Single-step supervision with SGD alone only partially absorbs real-world priors. See \S\ref{sec: dig}.}
        \label{fig:vis2}
\vspace{-16pt}
\end{figure}

\noindent\textbf{Inference Efficiency.}
As shown in Table~\ref{tab:efficiency}, Iris achieves favorable inference efficiency, running faster than DAv2~\citep{yang2024depthv2} and remaining substantially more efficient than multi-step diffusion-based methods such as Marigold~\citep{ke2024repurposing}.

\begin{table}[h]
\vspace{-6pt}
\caption{\textbf{Inference efficiency comparison.} Inference time (seconds) measured at $1536^2$ resolution on a NVIDIA A100 GPU.}
\label{tab:efficiency}
\vspace{-8pt}
\setlength{\tabcolsep}{3.8pt}
\renewcommand{\arraystretch}{1.2}
\centering
\footnotesize
\begin{tabular}{c|cccc}
\thickhline
\cellcolor[rgb]{0.92,0.92,0.92} Method & Marigold~\citep{ke2024repurposing} & Lotus~\citep{he2025lotus} & DAv2~\citep{yang2024depthv2} & \textbf{Iris} \\
\hline
\cellcolor[rgb]{0.92,0.92,0.92} Inference Time (s)~$\downarrow$ & 377.7 & 0.8 & 2.2 & 1.3 \\
\thickhline
\end{tabular}
\vspace{-12pt}
\end{table}

\subsection{Qualitative Results}
As seen in Fig.~\ref{fig:vis1}, Iris delivers depth maps with accurate metric scale and rich fine details and textures across diverse, challenging scenes, showcasing strong generalization and faithful detail modeling Please refer to the supplementary material for more visualizations.
\label{sec:qual_result}

\begin{table}[t]
\caption{\textbf{Ablation studies of essential components.} \emph{Deterministic} alone refers to single-step deterministic network. \emph{Deterministic + SGD} stands for simultaneous supervision on synthetic ground truth and real pseudo labels after single-stage. Note that \emph{two-stage} denotes Priors-to-Geometry pipeline. Collectively, \emph{Deterministic + two-stage + SGD + SGC} define our final Priors-to-Geometry Deterministic (PGD) framework. See \S\ref{sec: dig} for details.}
\centering
\vspace{-5pt}
\setlength{\tabcolsep}{0.2pt}
\renewcommand\arraystretch{1.05}
\resizebox{\columnwidth}{!}{%
\begin{tabular}{l|cc|cc}
\hline
\rowcolor[rgb]{0.92,0.92,0.92}
\rule{0pt}{1.6ex}\raisebox{-0.8ex}[1.6ex][1.3ex]{Components}
& \multicolumn{2}{c|}{KITTI (Outdoor)}
& \multicolumn{2}{c}{ScanNet (Indoor)} \\
\rowcolor[rgb]{0.92,0.92,0.92}
\rule{0pt}{1.6ex}
& AbsRel$\downarrow$ & $\delta_1\uparrow$
& AbsRel$\downarrow$ & $\delta_1\uparrow$ \\
\hline
(a)Stochastic      & 8.7 & 92.0  & 6.2 & 95.3 \\
\hline
(b)Deter.   & 8.3 & 92.8  & 5.8 & 96.0 \\
(c)Deter. + SGD & 7.9 & 93.2 & 5.6 & 96.2 \\
(d)Deter. + two-stage  & 8.2 & 93.0 & 5.7 & 96.0 \\
(e)Deter. + two-stage + vanilla distil  & 7.5 & 94.0 & 5.3 & 96.6 \\
(f)Deter. + two-stage + SGD & 7.4 & 94.2 & 5.2 & 96.9 \\
(g)\textcolor{red}{Deter. + two-stage + SGD + SGC} & \textbf{7.2} & \textbf{94.5} & \textbf{5.0} & \textbf{97.1} \\
\hline
\end{tabular}%
}
\vspace{-10pt}
\label{table:ab1}
\end{table}

\subsection{Diagnostic Experiments}
\label{sec: dig}
\noindent\textbf{Essential Components.} As shown in Table~\ref{table:ab1}, comparing (a) and (b), simply replacing the stochastic paradigm with a single-step deterministic paradigm yields gains, as also observed in recent studies~\citep{xu2025matters,he2025lotus}. Comparing entries (c) and (f), single-stage simultaneous supervision on synthetic ground truth and real pseudo labels yield limited gains due to gradient interference between low-frequency real priors and high-frequency synthetic cues; in contrast, the two-stage Priors-to-Geometry schedule fully exploits real-world priors, validating the necessity of stage decoupling. Comparing (e) and (f), vanilla distillation introduces teacher pseudo label artifacts and weakens detail modeling. Comparing (f) and (g), SGC further improves performance. Please refer to Fig.~\ref{fig:vis2} for visualization.

\noindent\textbf{Hyperparameters.} As can be seen in Table~\ref{table:ab2}, the reconstruction loss function further improves performance. Moreover, omitting the over-activation constraint in SGC causes stage-1 to over-amplify high-frequency content, which carries over to stage-2 and degrades accuracy.

\begin{table}[t]
\caption{\textbf{Ablation studies of hyperparameters.} $\alpha$, $\beta$, and $\gamma$ control the relative strengths of SGD, SGC, and the reconstruction loss, respectively. See \S\ref{sec: dig} for details.}
\centering
\vspace{-5pt}
\setlength{\tabcolsep}{1.2pt}
\renewcommand\arraystretch{1.05}
\resizebox{0.92\columnwidth}{!}{%
\begin{tabular}{c|cc|cc|cc}
\hline
\rowcolor[rgb]{0.92,0.92,0.92}
\rule{0pt}{1.8ex}\raisebox{-0.78ex}[1.8ex][1.3ex]{$(\alpha, \beta, \gamma)$}
& \multicolumn{2}{c|}{KITTI (Outdoor)}
& \multicolumn{2}{c|}{ETH3D (Various)}
& \multicolumn{2}{c}{ScanNet (Indoor)} \\
\rowcolor[rgb]{0.92,0.92,0.92}
\rule{0pt}{1.8ex}
& AbsRel$\downarrow$ & $\delta_1\uparrow$
& AbsRel$\downarrow$ & $\delta_1\uparrow$
& AbsRel$\downarrow$ & $\delta_1\uparrow$ \\
\hline
(0, 0, 0)   & 8.3 & 92.8  & 6.2 & 96.8 & 5.8 & 95.8 \\
\hline
(0.5, 0, 0) & 7.8 & 93.7  & 5.9 & 97.0 & 5.5 & 96.4 \\
(1, 0, 0)   & 7.5 & 94.1  & 5.7 & 97.2 & 5.3 & 96.7 \\
(0, 0, 1)   & 8.2 & 93.0  & 6.1 & 96.9 & 5.8 & 95.9 \\
(1, 0, 1)   & 7.4 & 94.2  & 5.6 & 97.4 & 5.2 & 96.9 \\
(1, 1, 1)   & 7.9 & 93.2  & 6.2 & 96.8 & 5.6 & 96.0 \\
(\textcolor{red}{1, 0.1, 1}) & \textbf{7.2} & \textbf{94.5} & \textbf{5.5} & \textbf{97.6} & \textbf{5.0} & \textbf{97.1} \\
\hline
\end{tabular}%
}
\vspace{-16pt}
\label{table:ab2}
\end{table}

%% file: sec/5_conclusion.tex
% \vspace{6pt}
\section{Conclusion}
In this paper, we present \textbf{Iris}, a Priors-to-Geometry Deterministic framework that injects real-world priors into the diffusion model for monocular depth estimation. Our two-stage schedule separates prior alignment at a high timestep, where Spectral-Gated Distillation transfers low-frequency real priors, from geometry refinement at a low timestep, where Spectral-Gated Consistency enforces high-frequency agreement under an over-activation constraint. Together with an auxiliary reconstruction constraint, this design preserves the backbone's fine-detail modeling capacity while stabilizing training under mixed synthetic-real supervision. Iris preserves fine details, generalizes strongly from synthetic to real scenes, and remains data-efficient, achieving significant improvements across diverse real-image benchmarks and outperforming both prior diffusion-based methods and large-scale deterministic feed-forward models.
% We hope this work can provide some insights for the community.

\noindent\textbf{Acknowledgement.} This work was supported by Zhejiang Provincial Natural Science Foundation of China (No. LR26F020002), the National Natural Science Foundation of China (No. 62472222, 62372405), Natural Science Foundation of Jiangsu Province (No. BK20240080), Fundamental Research Funds for the Central Universities (226-2025-00057), National Defense Science and Technology Industry Bureau Technology Infrastructure Project (JSZL2024606C001), and the State Key Laboratory of Brain Cognition and Brain-inspired Intelligence Technology (No. SKLBI-K2025004).

%% file: sec/X_suppl.tex
\clearpage
\setcounter{page}{1}
\setcounter{section}{0}
\setcounter{table}{0}
\setcounter{figure}{0}
\renewcommand{\thetable}{S\arabic{table}}
\renewcommand{\thefigure}{S\arabic{figure}}
\maketitlesupplementary
\appendix

\centerline{\textbf{SUMMARY OF THE APPENDIX}}
This appendix contains additional details for CVPR2026 submission, titled \textit{Iris: Bringing Real-World Priors into Diffusion Model for Monocular Depth Estimation}, which is organized as follows:
\begin{itemize}
    \item \S\ref{sec_app:discussion} discusses our limitations, directions of our future work, and societal impact.
    \item \S\ref{sec_app:mtl} introduces multi-task learning for zero-shot monocular depth and normal estimation with a single model.
    \item \S\ref{sec_app:quan} provides more quantitative results.
    \item \S\ref{sec_app:qua} provides more visualizations.
\end{itemize}

\section{Discussion and Outlook}
\label{sec_app:discussion}
\subsection{Limitation and Future Work}
Although Iris achieves the best overall performance among both traditional deterministic feed-forward models and diffusion-based methods, with particularly large improvements on outdoor and mixed benchmarks (\textit{e.g.,} KITTI~\citep{geiger2013vision}, ETH3D~\citep{schops2017multi}), the gains on certain indoor datasets (\textit{e.g.,} NYUv2~\citep{silberman2012indoor}) are more modest. We attribute this gap primarily to the distribution of our data source the SA-1B subset used for distillation is dominated by outdoor scenes and contains relatively few indoor scenes. As part of future work, we plan to incorporate more indoor real-image datasets and increase the scene diversity of our real-image supervision, in order to further enhance the scene generalization capability of Iris.

\subsection{Social Impact}
Iris is a generic monocular depth estimation framework that can be used as a building block in many 3D perception systems. By providing accurate and robust depth from a single RGB image, Iris can substantially lower the hardware and annotation cost of 3D perception. This has the potential to broaden access to 3D reconstruction in domains such as urban mapping, cultural heritage digitization, robotics, and AR/VR content creation. In robotics and autonomous navigation, improved monocular depth estimation can serve as a complementary signal to LiDAR or stereo sensors, providing redundancy in case of sensor degradation and enabling more affordable platforms that rely primarily on cameras.

However, using monocular depth in safety-critical settings such as autonomous driving also introduces risks. Rare but large depth errors, domain shift to unseen environments, or biases stemming from unbalanced training data may all lead to incorrect distance estimation and unsafe control decisions if the model is used as a primary sensor. In addition, the ability to recover dense 3D geometry from ordinary images may raise privacy concerns when applied to people or private spaces without consent.

\section{Multi-task Learning}
\label{sec_app:mtl}
\begin{figure*}[t]
    \centering
    \includegraphics[width=\textwidth]{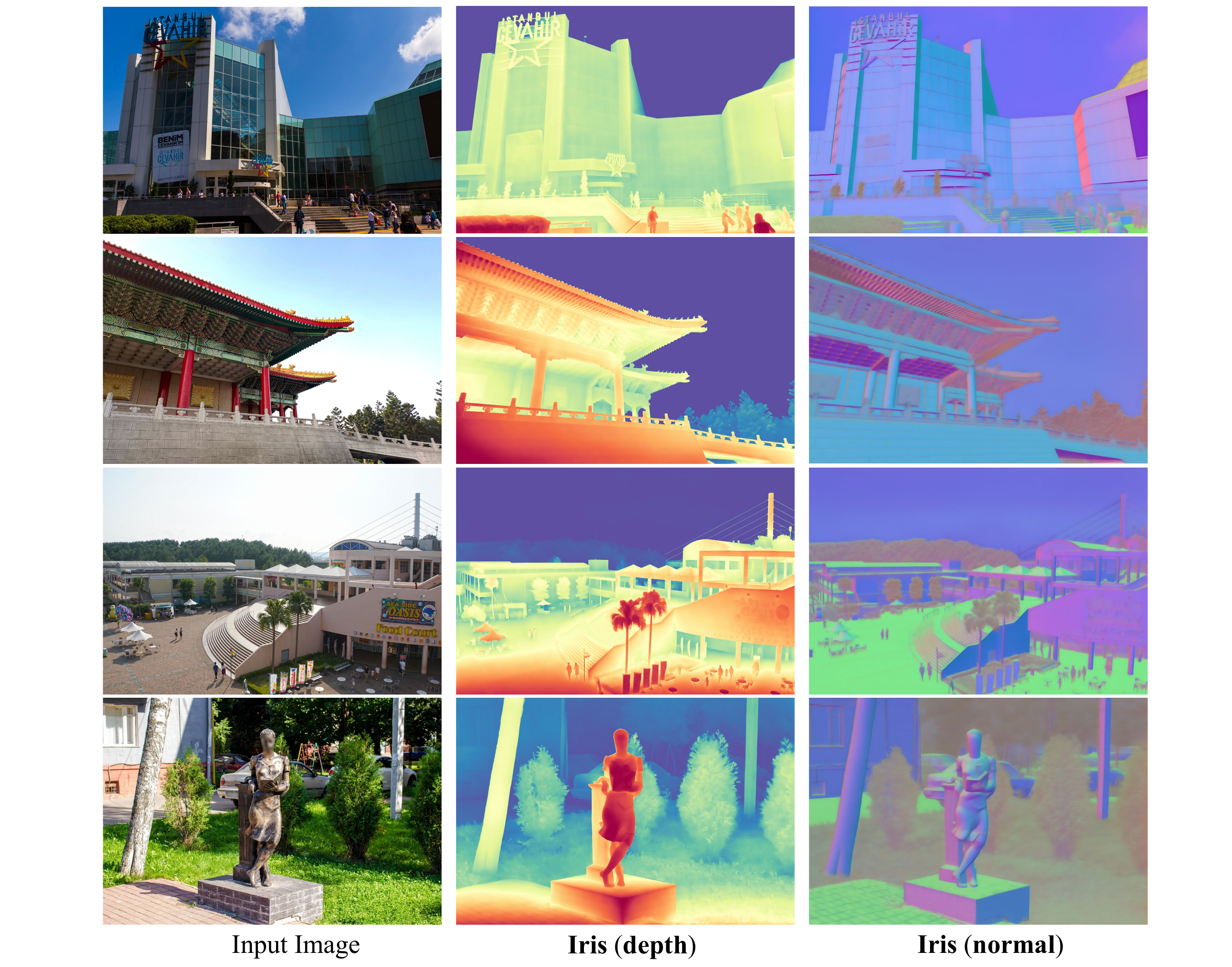}
    \vspace{-20pt}
    \caption{\textbf{Visualizations of Joint depth and normal estimation.} Iris enables simultaneous depth and normal estimation with \emph{fully shared parameters} by swapping the task switcher $s_y^{\text{depth}}$ and $s_y^{\text{normal}}$. See \S\ref{sec_app:mtl} for details.}
    \label{fig:vis4}
    \vspace{-6pt}
\end{figure*}

\begin{figure*}[t]
    \centering
    \includegraphics[width=\textwidth]{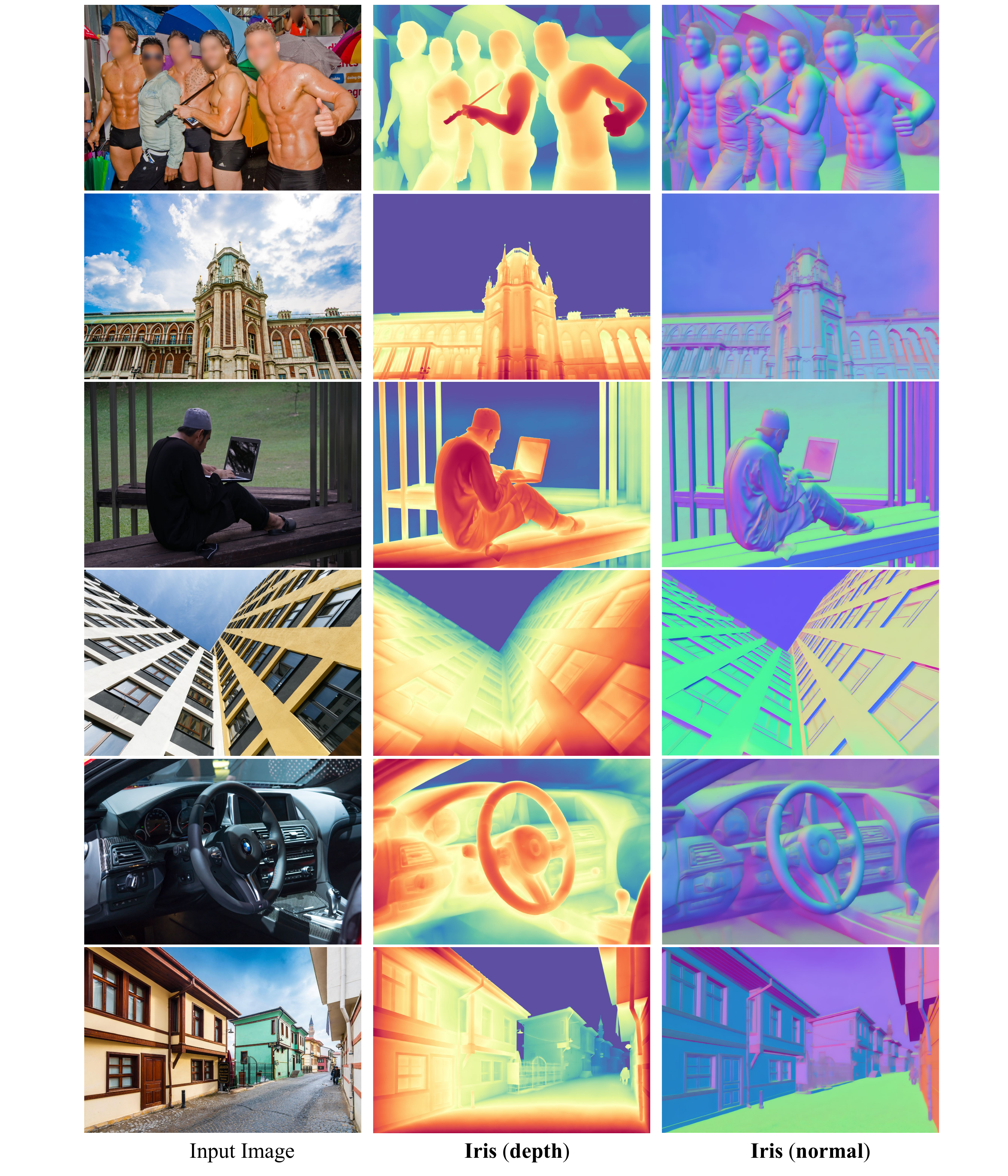}
    \vspace{-20pt}
    \caption{\textbf{Visualizations of Joint depth and normal estimation.} Iris enables simultaneous depth and normal estimation with \emph{fully shared parameters} by swapping the task switcher $s_y^{\text{depth}}$ and $s_y^{\text{normal}}$. See \S\ref{sec_app:mtl} for details.}
    \label{fig:vis5}
    \vspace{-6pt}
\end{figure*}

While the main paper primarily focuses on the depth estimation task, we demonstrate that \emph{simultaneous depth and normal estimation} can be achieved with \emph{fully shared parameters} in a single diffusion-based framework. This is realized through the integration of parameter sharing and task embedding injection. Let $s_x$ denotes the reconstruction task switcher (Eq.~\ref{eq:reconstruct}), and $s_y$ denotes the switcher for dense prediction. Throughout the main text, $s_y$ is tailored to depth estimation. However, In the context of simultaneous estimation, the dense prediction switcher $s_y$ takes values from the set $\{s_y^{\text{depth}}, s_y^{\text{normal}}\}$, allowing the model to adapt to the specific modality. During inference, the model can seamlessly transition between depth estimation and normal prediction solely by toggling the switcher $s_y$. See the Fig.~\ref{fig:vis5} and Fig.~\ref{fig:vis4} for visualizations.

\section{More Quantitative Results}
\label{sec_app:quan}
Table~\ref{tab:depth-da2k} shows the additional quantitative results on DA-2K, which is introducted by Depth Anything V2~\citep{yang2024depthv2}. On this benchmark, Iris outperforms all diffusion-based methods by large margins and narrows the gap to DepthAnything V2, which is trained on massive real-image corpora, demonstrating strong real-world generalization. 

\section{More Qualitative Results}
\label{sec_app:qua}
\begin{figure*}[t]
    \centering
    \includegraphics[width=\textwidth]{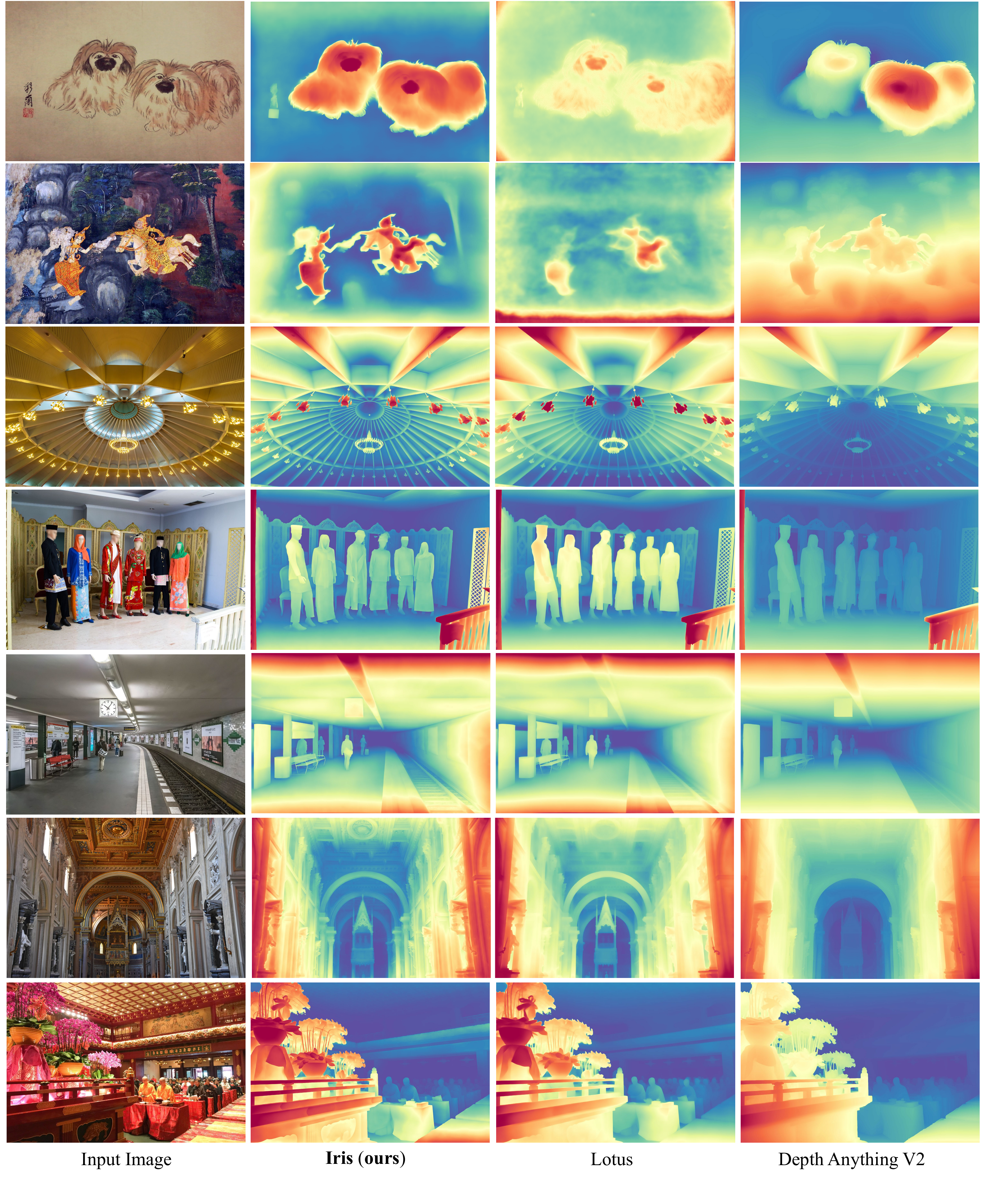}
    \vspace{-20pt}
    \caption{\textbf{More qualitative results on indoor scenes and paintings.} See \S\ref{sec_app:qua} for more details.}
    \label{fig:vis6}
    \vspace{-6pt}
\end{figure*}

\begin{figure*}[t]
    \centering
    \includegraphics[width=\textwidth]{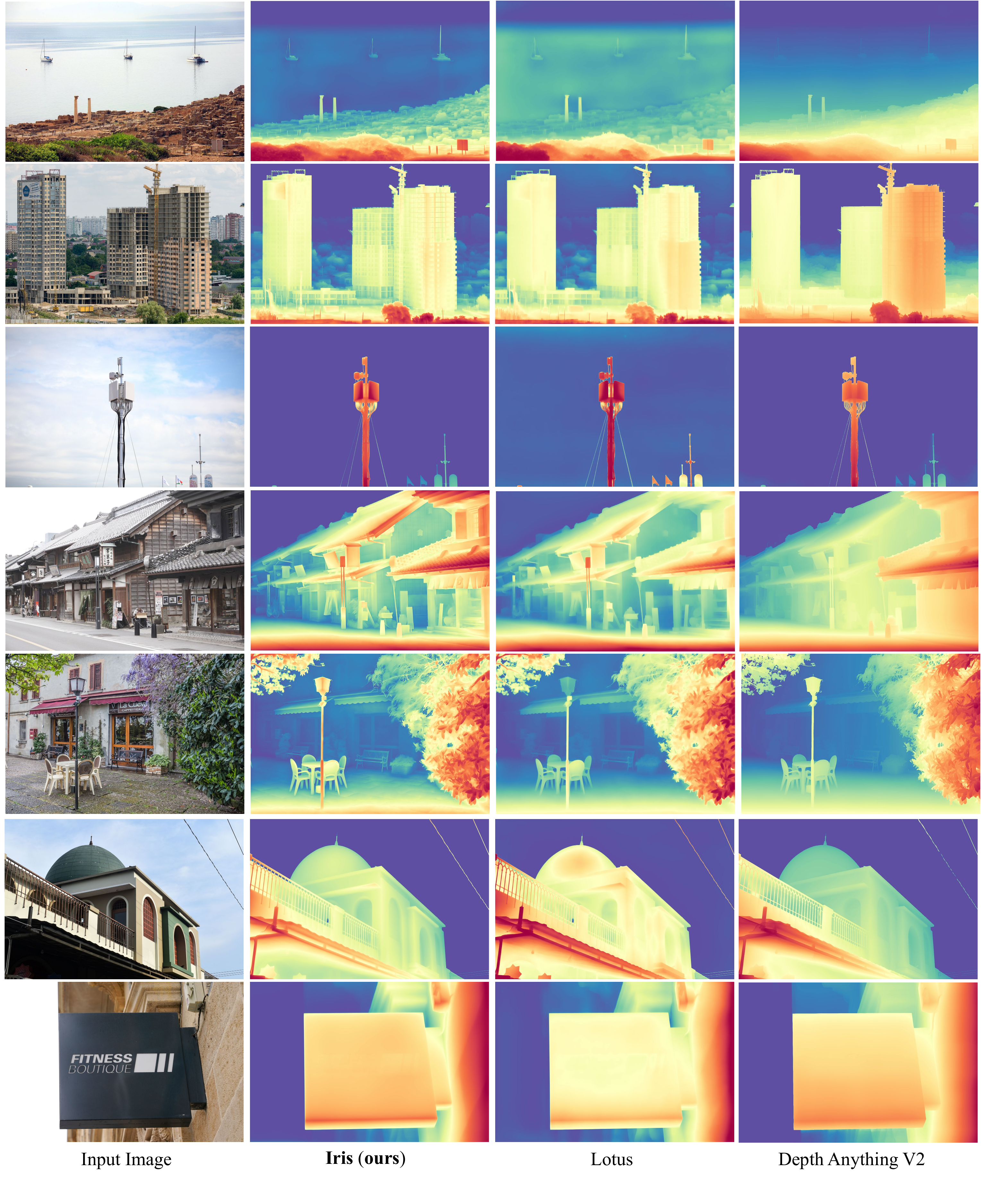}
    \vspace{-20pt}
    \caption{\textbf{More qualitative results on outdoor scenes.} See \S\ref{sec_app:qua} for more details.}
    \label{fig:vis7}
    \vspace{-20pt}
\end{figure*}

We provide additional qualitative results on indoor scenes and paintings in Fig.~\ref{fig:vis6}, and on outdoor scenes in Fig.~\ref{fig:vis7}. Since Lotus~\citep{he2025lotus} is trained only on synthetic datasets, it almost fails to produce meaningful depth on paintings (\textit{i.e.,} Fig.~\ref{fig:vis6} line 1-2). In contrast, Iris recovers plausible and detailed depth for these challenging artistic images. Across both indoor and outdoor scenes, Iris further demonstrates stronger scale awareness than Lotus, and delivers sharper object boundaries and richer fine-grained details than both Lotus and Depth Anything V2~\citep{yang2024depthv2}.

\setlength{\tabcolsep}{2.pt}
\begin{table*}[t]
% \vspace{-10mm}
% \scriptsize
\vspace{-3mm}
% \scriptsize
\caption{\textbf{Quantitative comparison on zero-shot affine-invariant depth estimation}.} 
\vspace{-6pt}
\resizebox{\textwidth}{!}{
\begin{tabular}{l|c|cc|cc|cc|cc|cc|c|c}
\toprule

\multirow{2}{*}{Method} & Training
& \multicolumn{2}{c|}{KITTI (Outdoor)} & \multicolumn{2}{c|}{NYUv2 (Indoor)} 
& \multicolumn{2}{c|}{ETH3D (Various)} & \multicolumn{2}{c|}{ScanNet (Indoor)} 
& \multicolumn{2}{c|}{\textcolor{black}{DIODE (Various)}} & \textbf{DA-2K}
& \textcolor{black}{Group Avg} \\
 & Data$\downarrow$
 & AbsRel~$\downarrow$ & $\delta_1$$\uparrow$
 & AbsRel~$\downarrow$ & $\delta_1$$\uparrow$
 & AbsRel~$\downarrow$ & $\delta_1$$\uparrow$
 & AbsRel~$\downarrow$ & $\delta_1$$\uparrow$
 & \textcolor{black}{AbsRel~$\downarrow$} & \textcolor{black}{$\delta_1$$\uparrow$} & Acc(\%) $\uparrow$
 &    \textcolor{black}{ Ranking}      \\
\midrule

DiverseDepth~\citep{yin2020diversedepth}
 & 320K
& 19.0 & 70.4
& 11.7 & 87.5
& 22.8 & 69.4
& 10.9 & 88.2
& \textcolor{black}{37.6} & \textcolor{black}{63.1}
& \textcolor{black}{79.3}   & 7.5            \\

MiDaS~\citep{ranftl2020towards}
 & 2M
& 23.6 & 63.0
& 11.1 & 88.5
& 18.4 & 75.2
& 12.1 & 84.6
& \textcolor{black}{33.2} & \textcolor{black}{71.5}
&  \textcolor{black}{80.6}    & 7.2             \\

LeRes~\citep{yin2021learning}
 & 354K
& 14.9 & 78.4
& 9.0 & 91.6
& 17.1 & 77.7
& 9.1 & 91.7
& \textcolor{black}{27.1} & \textcolor{black}{76.6}
& \textcolor{black}{81.1}     & 5.2             \\

Omnidata~\citep{eftekhar2021omnidata}
 & 12.2M
& 14.9 & 83.5
& 7.4 & 94.5
& 16.6 & 77.8
& 7.5 & 93.6
& \textcolor{black}{33.9} & \textcolor{black}{74.2}
& \textcolor{black}{76.8} & 5.0 \\

DPT~\citep{ranftl2021vision}
 & 1.4M
& 10.0 & 90.1
& 9.8 & 90.3
& \textbf{7.8} & \textbf{94.6}
& 8.2 & 93.4
& \textcolor{black}{\textbf{18.2}} & 75.8
&  \textcolor{black}{83.2}   & 3.5             \\

HDN~\citep{zhang2022hierarchical}
 & 300K
& 11.5 & 86.7
& 6.9 & 94.8
& 12.1 & 83.3
& 8.0 & 93.9
& \textcolor{black}{24.6} & \textcolor{black}{\textbf{78.0}}
& \textcolor{black}{85.7}   & 3.0            \\

\textcolor{black}{DepthAnything}~\citep{yang2024depth}  
& \textcolor{black}{62.6M}
& \textcolor{black}{7.6} & \textcolor{black}{\textbf{94.7}}
& \textcolor{black}{\textbf{4.3}} & \textcolor{black}{\textbf{98.1}}
& \textcolor{black}{12.7} & \textcolor{black}{88.2}
& \textcolor{black}{4.3} & \textcolor{black}{\textbf{98.1}}
& \textcolor{black}{26.0} & \textcolor{black}{75.9}
& \textcolor{black}{88.5}  & \textbf{1.9} \\

\textcolor{black}{DepthAnything V2}~\citep{yang2024depthv2}
& \textcolor{black}{62.6M}
& \textcolor{black}{\textbf{7.4}} & \textcolor{black}{94.6}
& \textcolor{black}{4.5} & \textcolor{black}{97.9}
& \textcolor{black}{13.1} & \textcolor{black}{86.5}
& \textcolor{black}{\textbf{4.2}} & \textcolor{black}{97.8}
& \textcolor{black}{26.5} & \textcolor{black}{73.4}
& \textcolor{black}{\textbf{97.1}} & 2.5 \\

\midrule

\textcolor{black}{Diffusion-E2E-FT$^{\star}$}~\citep{garcia2025fine}
 & \textcolor{black}{74K}
 & \textcolor{black}{9.6} & \textcolor{black}{92.1}
 & \textcolor{black}{5.4} & \textcolor{black}{96.5}
 & \textcolor{black}{6.4} & \textcolor{black}{95.9}
 & \textcolor{black}{5.8} & \textcolor{black}{96.5}
 & \textcolor{black}{30.3} & \textcolor{black}{\textbf{77.6}}
 & \textcolor{black}{83.6} & 4.3 \\

GeoWizard$^{\star}$~\citep{fu2024geowizard}
 & 280K
& 14.4 & 82.0
& 5.6 & 96.3
& 6.6 & 95.8
& 6.4 & 95.0
& \textcolor{black}{33.5} & \textcolor{black}{72.3}
&  \textcolor{black}{88.1}    & 6.8            \\

Marigold$_\text{(LCM)}$$^{\star}$~\citep{ke2024repurposing}
 & 74K
& 9.8 & 91.8
& 6.1 & 95.8
& 6.8 & 95.6
& 6.9 & 94.6
& \textcolor{black}{30.7} & \textcolor{black}{77.5}
&  \textcolor{black}{86.8}    & 6.5          \\

Marigold$^{\star}$~\citep{ke2024repurposing}
& 74K
& 9.9 & 91.6
& 5.5 & 96.4
& 6.5 & 95.9
& 6.4 & 95.2
& \textcolor{black}{30.8} & \textcolor{black}{77.3}
&  \textcolor{black}{85.6}    & 5.8          \\

\textcolor{black}{Lotus-D$^{\star}$}~\citep{he2025lotus}
& \textcolor{black}{59K}
& \textcolor{black}{8.1} & \textcolor{black}{93.1}
& \textcolor{black}{5.1} & \textcolor{black}{97.2}
& \textcolor{black}{6.1} & \textcolor{black}{97.0}
& \textcolor{black}{5.5} & \textcolor{black}{96.5}
& \textcolor{black}{22.8} & \textcolor{black}{73.8}
& \textcolor{black}{91.2}  & 2.7 \\

\textcolor{black}{Lotus-G$^{\star}$}~\citep{he2025lotus}
 & \textcolor{black}{59K}
 & \textcolor{black}{8.5} & \textcolor{black}{92.2}
 & \textcolor{black}{5.4} & \textcolor{black}{96.8}
 & \textcolor{black}{5.9} & \textcolor{black}{97.0}
 & \textcolor{black}{5.9} & \textcolor{black}{95.7}
 & \textcolor{black}{22.9} & \textcolor{black}{72.9}
 & \textcolor{black}{88.9} & 3.7 \\

GenPercept$^{\star}$~\citep{xu2025matters}     
 & 90K
& 7.8 & 93.5
& 5.9 & 96.7
& 9.4 & 96.1
& 6.4 & 96.1
& \textcolor{black}{\textbf{22.8}} & \textcolor{black}{74.0}
& \textcolor{black}{85.1}      & 4.6       \\

\textbf{Iris} (\textbf{ours})$^{\star}$ 
 & 59K + 100K
& \textbf{7.2} & \textbf{94.5}
& \textbf{4.9} & \textbf{97.4}
& \textbf{5.5} & \textbf{97.6}
& \textbf{5.0} & \textbf{97.1}
& \textcolor{black}{24.3} & \textcolor{black}{74.3}
& \textcolor{black}{\textbf{94.5}}   & \textbf{1.5}         \\

\bottomrule
\end{tabular}
}
\centering
\label{tab:depth-da2k}
\end{table*}

\clearpage

%% file: main.bib
@String(PAMI = {IEEE Trans. Pattern Anal. Mach. Intell.})

@String(CVPR= {IEEE Conf. Comput. Vis. Pattern Recog.})

@String(ICCV= {Int. Conf. Comput. Vis.})

@String(ECCV= {Eur. Conf. Comput. Vis.})

@String(NIPS= {Adv. Neural Inform. Process. Syst.})

@String(TOG= {ACM Trans. Graph.})

@String(TMM  = {IEEE Trans. Multimedia})

@String(ICASSP=	{ICASSP})

@String(ACCV  = {ACCV})

@String(ICLR = {Int. Conf. Learn. Represent.})

@String(PAMI  = {IEEE TPAMI})

@String(CVPR  = {CVPR})

@String(ICCV  = {ICCV})

@String(ECCV  = {ECCV})

@String(NIPS  = {NeurIPS})

@String(ICML  = {ICML})

@String(TOG   = {ACM TOG})

@String(TMM   =	{IEEE TMM})

@String(WACV  = {WACV})

@String(ICLR  = {ICLR})

@inproceedings{sohl2015deep,
  title={Deep unsupervised learning using nonequilibrium thermodynamics},
  author={Sohl-Dickstein, Jascha and Weiss, Eric and Maheswaranathan, Niru and Ganguli, Surya},
  booktitle=ICML,
  pages={2256--2265},
  year={2015},
}

@inproceedings{dhariwal2021diffusion,
  title={Diffusion models beat gans on image synthesis},
  author={Dhariwal, Prafulla and Nichol, Alexander},
  pages={8780--8794},
  booktitle=NIPS,
  year={2021}
}

@article{ho2022classifier,
  title={Classifier-free diffusion guidance},
  author={Ho, Jonathan and Salimans, Tim},
  journal={arXiv preprint arXiv:2207.12598},
  year={2022}
}

@inproceedings{kingma2021variational,
  title={Variational diffusion models},
  author={Kingma, Diederik and Salimans, Tim and Poole, Ben and Ho, Jonathan},
  booktitle=NIPS,
  year={2021}
}

@inproceedings{ho2020denoising,
  title={Denoising diffusion probabilistic models},
  author={Ho, Jonathan and Jain, Ajay and Abbeel, Pieter},
  booktitle=NIPS,
  pages={6840--6851},
  year={2020}
}

@article{song2020denoising,
  title={Denoising diffusion implicit models},
  author={Song, Jiaming and Meng, Chenlin and Ermon, Stefano},
  journal={arXiv preprint arXiv:2010.02502},
  year={2020}
}

@article{song2020score,
  title={Score-based generative modeling through stochastic differential equations},
  author={Song, Yang and Sohl-Dickstein, Jascha and Kingma, Diederik P and Kumar, Abhishek and Ermon, Stefano and Poole, Ben},
  journal={arXiv preprint arXiv:2011.13456},
  year={2020}
}

@inproceedings{rombach2022high,
  title={High-resolution image synthesis with latent diffusion models},
  author={Rombach, Robin and Blattmann, Andreas and Lorenz, Dominik and Esser, Patrick and Ommer, Bj{\"o}rn},
  booktitle=CVPR,
  pages={10684--10695},
  year={2022}
}

@article{nichol2021glide,
  title={Glide: Towards photorealistic image generation and editing with text-guided diffusion models},
  author={Nichol, Alex and Dhariwal, Prafulla and Ramesh, Aditya and Shyam, Pranav and Mishkin, Pamela and McGrew, Bob and Sutskever, Ilya and Chen, Mark},
  journal={arXiv preprint arXiv:2112.10741},
  year={2021}
}

@inproceedings{schuhmann2022laion,
  title={Laion-5b: An open large-scale dataset for training next generation image-text models},
  author={Schuhmann, Christoph and Beaumont, Romain and Vencu, Richard and Gordon, Cade and Wightman, Ross and Cherti, Mehdi and Coombes, Theo and Katta, Aarush and Mullis, Clayton and Wortsman, Mitchell and others},
  booktitle=NIPS,
  pages={25278--25294},
  year={2022}
}

@inproceedings{saxena2023surprising,
  title={The surprising effectiveness of diffusion models for optical flow and monocular depth estimation},
  author={Saxena, Saurabh and Herrmann, Charles and Hur, Junhwa and Kar, Abhishek and Norouzi, Mohammad and Sun, Deqing and Fleet, David J},
  booktitle=NIPS,
  pages={39443--39469},
  year={2023}
}

@inproceedings{luo2024flowdiffuser,
  title={Flowdiffuser: Advancing optical flow estimation with diffusion models},
  author={Luo, Ao and Li, Xin and Yang, Fan and Liu, Jiangyu and Fan, Haoqiang and Liu, Shuaicheng},
  booktitle=CVPR,
  pages={19167--19176},
  year={2024}
}

@inproceedings{li2023open,
  title={Open-vocabulary object segmentation with diffusion models},
  author={Li, Ziyi and Zhou, Qinye and Zhang, Xiaoyun and Zhang, Ya and Wang, Yanfeng and Xie, Weidi},
  booktitle=CVPR,
  pages={7667--7676},
  year={2023}
}

@inproceedings{xu2025matters,
  title={What matters when repurposing diffusion models for general dense perception tasks?},
  author={Xu, Guangkai and Ge, Yongtao and Liu, Mingyu and Fan, Chengxiang and Xie, Kangyang and Zhao, Zhiyue and Chen, Hao and Shen, Chunhua},
  booktitle=ICLR,
  year={2025}
}

@inproceedings{ke2024repurposing,
  title={Repurposing diffusion-based image generators for monocular depth estimation},
  author={Ke, Bingxin and Obukhov, Anton and Huang, Shengyu and Metzger, Nando and Daudt, Rodrigo Caye and Schindler, Konrad},
  booktitle=CVPR,
  pages={9492--9502},
  year={2024}
}

@inproceedings{fu2024geowizard,
  title={Geowizard: Unleashing the diffusion priors for 3d geometry estimation from a single image},
  author={Fu, Xiao and Yin, Wei and Hu, Mu and Wang, Kaixuan and Ma, Yuexin and Tan, Ping and Shen, Shaojie and Lin, Dahua and Long, Xiaoxiao},
  booktitle=ECCV,
  pages={241--258},
  year={2024},
}

@inproceedings{garcia2025fine,
  title={Fine-tuning image-conditional diffusion models is easier than you think},
  author={Garcia, Gonzalo Martin and Abou Zeid, Karim and Schmidt, Christian and De Geus, Daan and Hermans, Alexander and Leibe, Bastian},
  booktitle=WACV,
  pages={753--762},
  year={2025},
}

@inproceedings{he2025lotus,
  title={Lotus: Diffusion-based visual foundation model for high-quality dense prediction},
  author={He, Jing and Li, Haodong and Yin, Wei and Liang, Yixun and Li, Leheng and Zhou, Kaiqiang and Zhang, Hongbo and Liu, Bingbing and Chen, Ying-Cong},
  booktitle=ICLR,
  year={2025}
}

@inproceedings{bai2025fiffdepth,
  title={Fiffdepth: Feed-forward transformation of diffusion-based generators for detailed depth estimation},
  author={Bai, Yunpeng and Huang, Qixing},
  booktitle=ICCV,
  pages={6023--6033},
  year={2025}
}

@article{xu2025pixel,
  title={Pixel-Perfect Depth with Semantics-Prompted Diffusion Transformers},
  author={Xu, Gangwei and Lin, Haotong and Luo, Hongcheng and Wang, Xianqi and Yao, Jingfeng and Zhu, Lianghui and Pu, Yuechuan and Chi, Cheng and Sun, Haiyang and Wang, Bing and others},
  journal={arXiv preprint arXiv:2510.07316},
  year={2025}
}

@article{ye2024stablenormal,
  title={Stablenormal: Reducing diffusion variance for stable and sharp normal},
  author={Ye, Chongjie and Qiu, Lingteng and Gu, Xiaodong and Zuo, Qi and Wu, Yushuang and Dong, Zilong and Bo, Liefeng and Xiu, Yuliang and Han, Xiaoguang},
  journal={TOG},
  volume={43},
  number={6},
  pages={1--18},
  year={2024},
}

@inproceedings{eigen2014depth,
  title={Depth map prediction from a single image using a multi-scale deep network},
  author={Eigen, David and Puhrsch, Christian and Fergus, Rob},
  booktitle=NIPS,
  volume={27},
  year={2014}
}

@inproceedings{fu2018deep,
  title={Deep ordinal regression network for monocular depth estimation},
  author={Fu, Huan and Gong, Mingming and Wang, Chaohui and Batmanghelich, Kayhan and Tao, Dacheng},
  booktitle=CVPR,
  pages={2002--2011},
  year={2018}
}

@article{lee2019big,
  title={From big to small: Multi-scale local planar guidance for monocular depth estimation},
  author={Lee, Jin Han and Han, Myung-Kyu and Ko, Dong Wook and Suh, Il Hong},
  journal={arXiv preprint arXiv:1907.10326},
  year={2019}
}

@inproceedings{yuan2022neural,
  title={Neural window fully-connected crfs for monocular depth estimation},
  author={Yuan, Weihao and Gu, Xiaodong and Dai, Zuozhuo and Zhu, Siyu and Tan, Ping},
  booktitle=CVPR,
  pages={3916--3925},
  year={2022}
}

@inproceedings{li2018megadepth,
  title={Megadepth: Learning single-view depth prediction from internet photos},
  author={Li, Zhengqi and Snavely, Noah},
  booktitle=CVPR,
  pages={2041--2050},
  year={2018}
}

@article{yin2020diversedepth,
  title={Diversedepth: Affine-invariant depth prediction using diverse data},
  author={Yin, Wei and Wang, Xinlong and Shen, Chunhua and Liu, Yifan and Tian, Zhi and Xu, Songcen and Sun, Changming and Renyin, Dou},
  journal={arXiv preprint arXiv:2002.00569},
  year={2020}
}

@article{ranftl2020towards,
  title={Towards robust monocular depth estimation: Mixing datasets for zero-shot cross-dataset transfer},
  author={Ranftl, Ren{\'e} and Lasinger, Katrin and Hafner, David and Schindler, Konrad and Koltun, Vladlen},
  journal=PAMI,
  volume={44},
  number={3},
  pages={1623--1637},
  year={2020},
}

@inproceedings{eftekhar2021omnidata,
  title={Omnidata: A scalable pipeline for making multi-task mid-level vision datasets from 3d scans},
  author={Eftekhar, Ainaz and Sax, Alexander and Malik, Jitendra and Zamir, Amir},
  booktitle=ICCV,
  pages={10786--10796},
  year={2021}
}

@inproceedings{ranftl2021vision,
  title={Vision transformers for dense prediction},
  author={Ranftl, Ren{\'e} and Bochkovskiy, Alexey and Koltun, Vladlen},
  booktitle=ICCV,
  pages={12179--12188},
  year={2021}
}

@inproceedings{yang2024depth,
  title={Depth anything: Unleashing the power of large-scale unlabeled data},
  author={Yang, Lihe and Kang, Bingyi and Huang, Zilong and Xu, Xiaogang and Feng, Jiashi and Zhao, Hengshuang},
  booktitle=CVPR,
  pages={10371--10381},
  year={2024}
}

@inproceedings{yang2024depthv2,
  title={Depth anything v2},
  author={Yang, Lihe and Kang, Bingyi and Huang, Zilong and Zhao, Zhen and Xu, Xiaogang and Feng, Jiashi and Zhao, Hengshuang},
  booktitle=NIPS,
  pages={21875--21911},
  year={2024}
}

@inproceedings{yin2023metric3d,
  title={Metric3d: Towards zero-shot metric 3d prediction from a single image},
  author={Yin, Wei and Zhang, Chi and Chen, Hao and Cai, Zhipeng and Yu, Gang and Wang, Kaixuan and Chen, Xiaozhi and Shen, Chunhua},
  booktitle=ICCV,
  pages={9043--9053},
  year={2023}
}

@article{hu2024metric3d,
  title={Metric3d v2: A versatile monocular geometric foundation model for zero-shot metric depth and surface normal estimation},
  author={Hu, Mu and Yin, Wei and Zhang, Chi and Cai, Zhipeng and Long, Xiaoxiao and Chen, Hao and Wang, Kaixuan and Yu, Gang and Shen, Chunhua and Shen, Shaojie},
  journal=PAMI,
  year={2024},
  publisher={IEEE}
}

@inproceedings{roberts2021hypersim,
  title={Hypersim: A photorealistic synthetic dataset for holistic indoor scene understanding},
  author={Roberts, Mike and Ramapuram, Jason and Ranjan, Anurag and Kumar, Atulit and Bautista, Miguel Angel and Paczan, Nathan and Webb, Russ and Susskind, Joshua M},
  booktitle=ICCV,
  pages={10912--10922},
  year={2021}
}

@article{cabon2020virtual,
  title={Virtual kitti 2},
  author={Cabon, Yohann and Murray, Naila and Humenberger, Martin},
  journal={arXiv preprint arXiv:2001.10773},
  year={2020}
}

@inproceedings{silberman2012indoor,
  title={Indoor segmentation and support inference from rgbd images},
  author={Silberman, Nathan and Hoiem, Derek and Kohli, Pushmeet and Fergus, Rob},
  booktitle=ECCV,
  pages={746--760},
  year={2012},
}

@inproceedings{dai2017scannet,
  title={Scannet: Richly-annotated 3d reconstructions of indoor scenes},
  author={Dai, Angela and Chang, Angel X and Savva, Manolis and Halber, Maciej and Funkhouser, Thomas and Nie{\ss}ner, Matthias},
  booktitle=CVPR,
  pages={5828--5839},
  year={2017}
}

@article{geiger2013vision,
  title={Vision meets robotics: The kitti dataset},
  author={Geiger, Andreas and Lenz, Philip and Stiller, Christoph and Urtasun, Raquel},
  journal={The international journal of robotics research},
  volume={32},
  number={11},
  pages={1231--1237},
  year={2013},
}

@inproceedings{schops2017multi,
  title={A multi-view stereo benchmark with high-resolution images and multi-camera videos},
  author={Schops, Thomas and Schonberger, Johannes L and Galliani, Silvano and Sattler, Torsten and Schindler, Konrad and Pollefeys, Marc and Geiger, Andreas},
  booktitle=CVPR,
  pages={3260--3269},
  year={2017}
}

@article{vasiljevic2019diode,
  title={Diode: A dense indoor and outdoor depth dataset},
  author={Vasiljevic, Igor and Kolkin, Nick and Zhang, Shanyi and Luo, Ruotian and Wang, Haochen and Dai, Falcon Z and Daniele, Andrea F and Mostajabi, Mohammadreza and Basart, Steven and Walter, Matthew R and others},
  journal={arXiv preprint arXiv:1908.00463},
  year={2019}
}

@inproceedings{kirillov2023segment,
  title={Segment anything},
  author={Kirillov, Alexander and Mintun, Eric and Ravi, Nikhila and Mao, Hanzi and Rolland, Chloe and Gustafson, Laura and Xiao, Tete and Whitehead, Spencer and Berg, Alexander C and Lo, Wan-Yen and others},
  booktitle=ICCV,
  pages={4015--4026},
  year={2023}
}

@inproceedings{ronneberger2015u,
  title={U-net: Convolutional networks for biomedical image segmentation},
  author={Ronneberger, Olaf and Fischer, Philipp and Brox, Thomas},
  booktitle=MICCAI,
  pages={234--241},
  year={2015},
}

@inproceedings{yang2023diffusion,
  title={Diffusion model as representation learner},
  author={Yang, Xingyi and Wang, Xinchao},
  booktitle=ICCV,
  pages={18938--18949},
  year={2023}
}

@inproceedings{zhao2023unleashing,
  title={Unleashing text-to-image diffusion models for visual perception},
  author={Zhao, Wenliang and Rao, Yongming and Liu, Zuyan and Liu, Benlin and Zhou, Jie and Lu, Jiwen},
  booktitle=ICCV,
  pages={5729--5739},
  year={2023}
}

@article{zhai2023investigating,
  title={Investigating the catastrophic forgetting in multimodal large language models},
  author={Zhai, Yuexiang and Tong, Shengbang and Li, Xiao and Cai, Mu and Qu, Qing and Lee, Yong Jae and Ma, Yi},
  journal={arXiv preprint arXiv:2309.10313},
  year={2023}
}

@inproceedings{wang2023sparsenerf,
  title={Sparsenerf: Distilling depth ranking for few-shot novel view synthesis},
  author={Wang, Guangcong and Chen, Zhaoxi and Loy, Chen Change and Liu, Ziwei},
  booktitle=ICCV,
  pages={9065--9076},
  year={2023}
}

@inproceedings{zhang2023adding,
  title={Adding conditional control to text-to-image diffusion models},
  author={Zhang, Lvmin and Rao, Anyi and Agrawala, Maneesh},
  booktitle=ICCV,
  pages={3836--3847},
  year={2023}
}

@inproceedings{hu2023planning,
  title={Planning-oriented autonomous driving},
  author={Hu, Yihan and Yang, Jiazhi and Chen, Li and Li, Keyu and Sima, Chonghao and Zhu, Xizhou and Chai, Siqi and Du, Senyao and Lin, Tianwei and Wang, Wenhai and others},
  booktitle=CVPR,
  pages={17853--17862},
  year={2023}
}

@inproceedings{yin2021learning,
  title={Learning to recover 3d scene shape from a single image},
  author={Yin, Wei and Zhang, Jianming and Wang, Oliver and Niklaus, Simon and Mai, Long and Chen, Simon and Shen, Chunhua},
  booktitle=CVPR,
  pages={204--213},
  year={2021}
}

@inproceedings{zhang2022hierarchical,
  title={Hierarchical normalization for robust monocular depth estimation},
  author={Zhang, Chi and Yin, Wei and Wang, Billzb and Yu, Gang and Fu, Bin and Shen, Chunhua},
  booktitle=NIPS,
  pages={14128--14139},
  year={2022}
}

@inproceedings{zavadski2024primedepth,
  title={Primedepth: Efficient monocular depth estimation with a stable diffusion preimage},
  author={Zavadski, Denis and Kal{\v{s}}an, Damjan and Rother, Carsten},
  booktitle=ACCV,
  pages={922--940},
  year={2024}
}

@inproceedings{cai2025cycle,
  title={Cycle-consistent learning for joint layout-to-image generation and object detection},
  author={Cai, Xinhao and Lai, Qiuxia and Pei, Gensheng and Shu, Xiangbo and Yao, Yazhou and Wang, Wenguan},
  booktitle=ICCV,
  pages={6797--6807},
  year={2025}
}

@inproceedings{cai2026unbiased,
  title={Unbiased Object Detection Beyond Frequency with Visually Prompted Image Synthesis},
  author={Cai, Xinhao and Li, Liulei and Pei, Gensheng and Chen, Tao and Pan, Jinshan and Yao, Yazhou and Wang, Wenguan},
  booktitle=ICLR,
  year={2026}
}

@article{Yin2025,
  title={Semi-Supervised Semantic Segmentation With Multi-Constraint Consistency Learning}, 
  author={Yin, Jianjian and Chen, Tao and Pei, Gensheng and Liu, Huafeng and Yao, Yazhou and Nie, Liqiang and Hua, Xiansheng},
  journal=TMM, 
  year={2025},
  volume={27},
  pages={6449-6461},
  }

@inproceedings{yin2025uncertainty,
  title={Uncertainty-participation context consistency learning for semi-supervised semantic segmentation},
  author={Yin, Jianjian and Chen, Yi and Zheng, Zhichao and Zhou, Junsheng and Gu, Yanhui},
  booktitle=ICASSP,
  pages={1--5},
  year={2025},
}

@inproceedings{zhou2025unialign,
  title={Unialign: Scaling multimodal alignment within one unified model},
  author={Zhou, Bo and Li, Liulei and Wang, Yujia and Liu, Huafeng and Yao, Yazhou and Wang, Wenguan},
  booktitle=CVPR,
  pages={29644--29655},
  year={2025}
}

@inproceedings{CA2C,
    author    = {Sheng, Mengmeng and Sun, Zeren and Zhou, Tianfei and Shu, Xiangbo and Pan, Jinshan and Yao, Yazhou},
    title     = {CA2C: A Prior-Knowledge-Free Approach for Robust Label Noise Learning via Asymmetric Co-learning and Co-training},
    booktitle = ICCV,
    year      = {2025},
    pages     = {901-911}
}

@inproceedings{sheng2024foster,
  title={Foster adaptivity and balance in learning with noisy labels},
  author={Sheng, Mengmeng and Sun, Zeren and Chen, Tao and Pang, Shuchao and Wang, Yucheng and Yao, Yazhou},
  booktitle=ECCV,
  pages={217--235},
  year={2024}
}

@inproceedings{deng2009imagenet,
  title={Imagenet: A large-scale hierarchical image database},
  author={Deng, Jia and Dong, Wei and Socher, Richard and Li, Li-Jia and Li, Kai and Fei-Fei, Li},
  booktitle=CVPR,
  pages={248--255},
  year={2009},
}
